\documentclass{article}
\pdfoutput=1

\usepackage[narrow]{arxiv}

\usepackage[utf8]{inputenc} 
\usepackage[T1]{fontenc}    
\usepackage[hypertexnames=false]{hyperref}       
\usepackage{url}            
\usepackage{booktabs}       
\usepackage{amsfonts}       
\usepackage{nicefrac}       
\usepackage{microtype}      

\usepackage{lipsum}         
\usepackage{graphicx}
\usepackage{natbib}
\usepackage{doi}

\usepackage{mathtools,amsthm,amssymb,thmtools}
\usepackage{algorithm}
\usepackage{algpseudocodex}
\usepackage{xcolor}
\usepackage{etoolbox}
\usepackage{todonotes}
\usepackage{environ}
\usepackage{cleveref}
\usepackage{crossreftools}
\usepackage{suffix}
\usepackage{nicematrix}
\usepackage{enumitem}
\usepackage{expl3}
\usepackage{makecell}
\usepackage{bm}

\hypersetup{
  colorlinks=true,
  linkcolor=red!75!black,
  citecolor=blue!50!black
}

\title{SGD with Adaptive Preconditioning: Unified Analysis and Momentum Acceleration}

\newif\ifuniqueAffiliation
\uniqueAffiliationtrue

\ifuniqueAffiliation 
\author{
  Dmitry Kovalev \\
  Yandex Research \\
  \texttt{dakovalev1@gmail.com}
}
\else
\usepackage{authblk}

\author[1]{%
  Dmitry Kovalev\thanks{\texttt{dakovalev1@gmail.com}}%
}
\affil[1]{Yandex Research}
\fi
\renewcommand{\headeright}{}

\newcommand{\argordot}[2]{\ifblank{#1}{#2\cdot#2}{#1}}
\DeclarePairedDelimiterXPP{\gennormdel}[3]{}\lVert\rVert{\ifblank{#2}{}{_#2}\ifblank{#3}{}{^{#3}}}{\argordot{#1}{\,}}

\NewDocumentCommand{\gennorm}{mmsm}{\IfBooleanTF{#3}{\gennormdel*{#4}{#1}{#2}}{\gennormdel{#4}{#1}{#2}}}

\newcommand{\norm}{\gennorm{}{}}
\newcommand{\sqn}{\gennorm{}{2}}
\newcommand{\normi}{\gennorm{\infty}{}}

\DeclarePairedDelimiter{\abs}\lvert\rvert

\providecommand{\given}{}
\DeclarePairedDelimiterX{\set}[1]\{\}{\renewcommand{\given}{:}#1}

\DeclarePairedDelimiterXPP{\Probability}[2]{\mathbb{P}\ifblank{#1}{}{_{#1}}}(){}{\renewcommand{\given}{\nonscript\:\delimsize\vert\nonscript\:}#2}
\DeclarePairedDelimiterXPP{\Expectation}[2]{\mathbb{E}\ifblank{#1}{}{_{#1}}}[]{}{\renewcommand{\given}{\nonscript\:\delimsize\vert\nonscript\:}#2}

\NewDocumentCommand{\E}{sO{}m}{\IfBooleanTF{#1}{\Expectation*{#2}{#3}}{\Expectation{#2}{#3}}}
\RenewDocumentCommand{\P}{sO{}m}{\IfBooleanTF{#1}{\Probability*{#2}{#3}}{\Probability{#2}{#3}}}

\def\<#1,#2>{\langle \argordot{#1}{},\argordot{#2}{}\rangle}
\WithSuffix\def\<*#1,#2>{\left\langle\argordot{#1}{},\argordot{#2}{}\right\rangle}

\def\mi[#1]{\hspace{#1em}&\hspace{-#1em}}

\DeclarePairedDelimiterX{\braround}[1](){\argordot{#1}{}}
\DeclarePairedDelimiterX{\brasquare}[1][]{\argordot{#1}{}}

\newcommand{\trace}{\operatorname{tr}\braround}
\newcommand{\diag}{\operatorname{diag}\braround}

\renewcommand{\dim}{\operatorname{dim}\braround}

\NewDocumentCommand{\proj}{sO{}m}{\operatorname{proj}\ifblank{#2}{}{_{#2}}\IfBooleanTF{#1}{\braround*{#3}}{\braround{#3}}}
\NewDocumentCommand{\prox}{sO{}m}{\operatorname{prox}\ifblank{#2}{}{_{#2}}\IfBooleanTF{#1}{\braround*{#3}}{\braround{#3}}}

\newcommand{\tsum}{{\textstyle\sum}}
\newcommand{\targmin}{{\textstyle\argmin}}

\DeclareMathOperator*{\argmin}{arg\,min}

\newcommand{\adj}{^*}
\newcommand{\T}{^\top}

\newcommand{\R}{\mathbb{R}}
\newcommand{\sL}{\mathbb{L}}
\newcommand{\Sym}{\mathbb{S}}
\newcommand{\Symp}{\Sym_+}
\newcommand{\Sympp}{\Sym_{++}}

\newcommand{\mSigma}{\mathbf{\Sigma}}

\newcommand{\lmax}{\lambda_{\max}}
\newcommand{\lmin}{\lambda_{\min}}

\newcommand{\smax}{\sigma_{\max}}

\newcommand{\clc}{,\ldots,}

\newcommand{\cA}{\mathcal{A}}
\newcommand{\cB}{\mathcal{B}}
\newcommand{\cC}{\mathcal{C}}
\newcommand{\cD}{\mathcal{D}}

\newcommand{\cH}{\mathcal{H}}

\newcommand{\cO}{\mathcal{O}}

\newcommand{\cQ}{\mathcal{Q}}
\newcommand{\cR}{\mathcal{R}}
\newcommand{\cS}{\mathcal{S}}

\newcommand{\cX}{\mathcal{X}}

\newcommand{\mA}{\mathbf{A}}
\newcommand{\mB}{\mathbf{B}}

\newcommand{\mG}{\mathbf{G}}
\newcommand{\mH}{\mathbf{H}}
\newcommand{\mI}{\mathbf{I}}

\newcommand{\mL}{\mathbf{L}}

\newcommand{\mN}{\mathbf{N}}
\newcommand{\mO}{\mathbf{O}}
\newcommand{\mP}{\mathbf{P}}
\newcommand{\mQ}{\mathbf{Q}}
\newcommand{\mR}{\mathbf{R}}
\newcommand{\mS}{\mathbf{S}}

\newcommand{\mX}{\mathbf{X}}

\allowdisplaybreaks[4]

\makeatletter
\newcommand{\customtheorem}[2]{%
  \newtheorem{inner-#1}{#2}%
  \crefalias{inner-#1}{#1}%
  \NewDocumentEnvironment{#1}{O{}D<>{}}{%
    \csname inner-#1\endcsname[label={##2},name={%
        ##1%
        \@ifundefined{r@proof:##2}{}{%
          \ifblank{##1}{}{, }\hyperref[proof:##2]{$\downarrow$}%
        }%
    }]%
  }{%
    \csname endinner-#1\endcsname%
  }%
}%
\makeatother

\customtheorem{theorem}{Theorem}
\customtheorem{lemma}{Lemma}

\newcommand{\proofsubsection}[1]{
  \subsection{Proof of \Cref{#1}}\label{proof:#1}
}
\newcommand{\proofsubsubsection}[1]{
  \subsubsection{Proof of \Cref{#1}}\label{proof:#1}
}

\setlist{topsep=0pt,partopsep=0pt}

\newtheorem{assumption}{Assumption}

\newtheorem{definition}{Definition}

\crefname{assumption}{assumption}{assumptions}
\crefname{problem}{problem}{problems}
\crefname{condition}{condition}{conditions}
\crefname{property}{property}{properties}
\creflabelformat{problem}{(#2#1#3)}

\newcounter{aequation}
\NewEnviron{aequation}{\refstepcounter{aequation}$$\BODY\eqno{\text{(A\theaequation)}}$$}
\crefname{aequation}{assumption}{assumptions}
\creflabelformat{aequation}{(#2A#1#3)}

\newcounter{question}
\newenvironment{question}{%
  \par\centering\refstepcounter{question}\em%
  \everypar={{\setbox0=\lastbox}\textnormal{\textbf{Q\arabic{question}.~}}\everypar={}}%
}{%
  \par%
  \ignorespacesafterend%
}%
\crefname{question}{question}{questions}
\creflabelformat{question}{#2#1#3}

\makeatletter
\newcommand{\linkdest}[1]{\Hy@raisedlink{\hypertarget{#1}{}}}
\makeatother
\ExplSyntaxOn
\seq_new:N \g_desc_seq
\seq_new:N \g_desc_nodup_seq
\int_new:N \g_index_int
\int_gset:Nn \g_index_int{1}
\newcommand{\at}[1]{
  \legacy_if:nF { measuring@ }{
    \seq_gput_right:Nn \g_desc_seq {#1}
    \linkdest{label\int_use:N\g_index_int\seq_count:N\g_desc_seq}
  }
  \overset{
    \text{
      (\hyperlink{desc\int_use:N\g_index_int\seq_count:N\g_desc_seq}{\int_to_alph:n{\seq_count:N\g_desc_seq}})
    }
  }
}
\newcommand{\annotate}{
  \seq_gset_eq:NN\g_desc_nodup_seq\g_desc_seq
  \seq_gremove_duplicates:N\g_desc_nodup_seq
  \seq_map_indexed_inline:Nn \g_desc_nodup_seq {
    \seq_clear_new:N \l_labels_seq
    \seq_map_indexed_inline:Nn \g_desc_seq{
      \tl_gset:Nx\l_tmpa_tl{\seq_item:Nn\g_desc_seq{####1}}
      \tl_gset:Nx\l_tmpb_tl{\seq_item:Nn\g_desc_nodup_seq{##1}}
      \tl_if_eq:NNT\l_tmpa_tl\l_tmpb_tl{
        \seq_put_right:Nn\l_labels_seq{
          \linkdest{desc\int_use:N\g_index_int####1}
          (\hyperlink{label\int_use:N\g_index_int####1}{\int_to_alph:n{####1}})
        }
      }
    }
    \seq_map_indexed_inline:Nn \l_labels_seq{
      \seq_item:Nn \l_labels_seq{####1}
      \int_compare:nNnT{\seq_count:N\l_labels_seq}>{(####1)+1}{,~}
      \int_compare:nNnT{\seq_count:N\l_labels_seq}<{(####1)+1}{~}
      \int_compare:nNnT{\seq_count:N\l_labels_seq}={(####1)+1}{~and~}
    }
    \seq_item:Nn\g_desc_nodup_seq{##1}
    \int_compare:nNnF{##1}={\seq_count:N\g_desc_nodup_seq}{;~}
  }
  \seq_gclear:N\g_desc_seq
  \int_gadd:Nn\g_index_int{1}
}
\ExplSyntaxOff

\begin{document}
\maketitle

\begin{abstract}
  In this paper, we revisit stochastic gradient descent (SGD) with AdaGrad-type preconditioning. Our contributions are twofold. First, we develop a unified convergence analysis of SGD with adaptive preconditioning under anisotropic or matrix smoothness and noise assumptions. This allows us to recover state-of-the-art convergence results for several popular adaptive gradient methods, including AdaGrad-Norm, AdaGrad, and ASGO/One-sided Shampoo. In addition, we establish the fundamental connection between two recently proposed algorithms, Scion and DASGO, and provide the first theoretical guarantees for the latter. Second, we show that the convergence of methods like AdaGrad and DASGO can be provably accelerated beyond the best-known rates using Nesterov momentum. Consequently, we obtain the first theoretical justification that AdaGrad-type algorithms can simultaneously benefit from both diagonal preconditioning and momentum, which may provide an ultimate explanation for the practical efficiency of Adam.
\end{abstract}


\newcommand{\normt}{\gennorm{\mathrm{tr}}{}}
\newcommand{\sqnt}{\gennorm{\mathrm{tr}}{2}}
\newcommand{\norms}{\gennorm{\mathrm{op}}{}}
\newcommand{\sqns}{\gennorm{\mathrm{op}}{2}}
\newcommand{\ox}{\overline{x}}
\newcommand{\dist}{\cR\braround}
\renewcommand{\d}{\mathrm{d}}
\newcommand{\tint}{{\textstyle\int}}

\section{Introduction}

The optimization community has shown strong interest in adaptive stochastic gradient optimization methods over recent years \citep{duchi2011adaptive,tieleman2012lecture,kingma2014adam,gupta2018shampoo,reddi2019convergence} due to their applications in deep learning \citep{lecun2015deep}. This research direction has notably led to the development of Adam \citep{kingma2014adam} and AdamW \citep{loshchilov2017decoupled}, algorithms with remarkable performance in training deep neural networks. Unfortunately, despite almost a decade of research, these algorithms continue to be the preferred choice for most deep learning tasks, particularly in the training of large language models \citep{achiam2023gpt,liu2024deepseek,grattafiori2024llama,team2023gemini}. The lack of worthy contenders to Adam and AdamW may be attributed to insufficient theoretical understanding of adaptive optimization algorithms. Therefore, the primary objective of this paper is to enhance the theoretical comprehension of this research area.
Formally speaking, we consider the following optimization problem:
\begin{equation}\label[problem]{eq:main}
  \min_{x \in \cX} f(x),
\end{equation}
where $\cX$ is a finite-dimensional Euclidean space, and $f(x)\colon \cX \to \R$ is a continuous convex\footnote{We discuss the justification for using the convexity assumption in \Cref{sec:convex}.} objective function. We assume that \cref{eq:main} has a solution $x^* \in \cX$.

\subsection{Baseline Algorithm: AdaGrad}

The starting point for the development of Adam was the gradient descent (GD) with the AdaGrad-Norm stepsizes \citep{streeter2010less}. Given the parameter $\eta > 0$ and the past gradients $g_i \in \partial f(x_i)$ for $i=0,\ldots,k$, this algorithm performs the following update:
\begin{equation}\label{eq:adagrad_norm}
  x_{k+1} = x_k - \eta_k g_k,\quad \text{where}\quad
  \eta_k = \tfrac{\eta}{\sqrt{\sum_{i=0}^k\sqn{g_i}}}.
\end{equation}
It is well known that AdaGrad-Norm can achieve the convergence rate $\cO(1/K)$ of GD with fixed stepsizes for smooth functions with Lipschitz-continuous gradients and the rate $\cO(1/\sqrt{K})$ of GD with diminishing step sizes for non-smooth Lipschitz functions or when only stochastic gradients are available \citep{orabona2023normalized,li2019convergence,levy2018online}. However, the main benefit of this algorithm is that it can achieve both rates with the single parameter choice $\eta \propto \norm{x^*}$. In other words, it can adapt to the level of smoothness and gradient noise of the function $f(x)$, which is called ``universality'' \citep{nesterov2015universal}. Furthermore, \citet{duchi2011adaptive,mcmahan2010adaptive} proposed the AdaGrad method, which performs a coordinate-wise variant of the update~\eqref{eq:adagrad_norm}, aiming to exploit the potential sparsity of the gradients $g_k$. Although they provided a limited theoretical justification for the benefits of coordinate-wise updates compared to scalar stepsizes~\eqref{eq:adagrad_norm}, AdaGrad and its modifications, such as RMSProp \citep{tieleman2012lecture} and Adam, have proven to be highly efficient in practice.

\subsection{Adaptive Gradient Methods with Structured Preconditioning}\label{sec:intro:precond}

Motivated by the success of AdaGrad, many adaptive optimization algorithms has been developed that fall into the category of gradient methods with preconditioning. Such algorithms use the update rule of the form
\begin{equation}\label{eq:gd}
  x_{k+1} = \argmin_{x \in \cX} \<g_k,x> + \tfrac{1}{2}\sqn{x - x_k}_{\mH_k^{-1}},
\end{equation}
where $\mH_k \in \Sympp$ is a symmetric positive definite preconditioning operator $\cX \to \cX$. Besides AdaGrad, which uses a diagonal preconditioning matrix, notable examples of such algorithms include Shampoo \citep{gupta2018shampoo} and its theoretically streamlined variants: One-sided Shampoo \citep{xie2025structured} and ASGO \citep{an2025asgo}. Motivated by the structure of neural networks, these algorithms are specifically designed for optimizing the function $f(X)\colon \R^{m\times n} \to \R$ of an $m\times n$ matrix argument and use preconditioners that respect the function's structure. In particular, One-sided Shampoo and ASGO use the preconditioner $\mH_k\colon G \mapsto (\sum_{i=0}^{k}G_iG_i\T)^{-1/2}G$, where $G \in \R^{m\times n}$ and $G_i \in \partial f(X_i)$. Overall, the practical performance of Shampoo and its Adam-like modification, SOAP \citep{vyas2024soap}, is comparable to that of Adam and sometimes exceeds it.

Here, we come to the following issue: every time an adaptive preconditioned gradient method is developed, one has to provide a separate convergence proof, even though the update rules in such algorithms, as well as the convergence proofs, often have a similar structure. Consequently, we arrive to the following question:
\begin{question}\label{q1}
  Can we develop a unified convergence analysis that would cover most existing adaptive preconditioned gradient methods, including  AdaGrad, Shampoo, ASGO, etc.?
\end{question}
A positive answer to this question was partially provided by the unified approach of \citet{gupta2017unified}, who showed that the preconditioner operator $\mH_k$ can be defined as a solution to a certain optimization problem over a linear subspace of self-adjoint operators $\cH \subset \Sym$. For instance, the update rule for AdaGrad-Norm and AdaGrad can be obtained by choosing $\cH$ to be the space of multiples of the identity and the space of diagonal operators, respectively. Unfortunately, the unified approach of \citet{gupta2017unified} has major flaws: it still requires separate convergence proofs for different algorithms, provides convergence guarantees only for non-smooth functions, and offers no explanation for the benefits of using general preconditioning operators.

\subsection{Matrix Smoothness and Acceleration}

\textbf{Matrix smoothness.}
In an attempt to find a theoretical justification for the success of adaptive preconditioned gradient methods, a considerable amount of recent research has focused on developing theoretical analyses of such methods under the assumption that the function smoothness, as well as the gradient noise level, is measured in terms of the weighted Euclidean norm $\norm{}_{\mB}$, where $\mB \in \Sympp$ is a self-adjoint positive definite operator. For instance, \citet{liu2024adagrad,jiang2024convergence} provided an analysis of AdaGrad under anisotropic smoothness, i.e., in the case of the diagonal operator $\mB\colon x\mapsto \bm{b}\odot x$, where $\bm{b},x \in \R^d$. When the vector $\bm{b}$ is sparse, they managed to prove substantially better theoretical convergence guarantees for AdaGrad compared to AdaGrad-Norm, thus obtaining theoretical justification for the practical benefits of diagonal preconditioning. Similarly, \citet{an2025asgo,xie2025structured} considered the matrix smoothness, i.e., the case where the operator $\mB\colon X \mapsto B X$, where the matrix $B \in \R^{m\times m}$ is symmetric and positive definite, and $X \in \R^{m\times n}$. This allowed them to theoretically justify the practical success of Shampoo-like algorithms. However, \Cref{q1} discussed above is relevant here: a separate convergence proof is required for each algorithm, even though they share many similarities.

\textbf{Momentum acceleration.} Besides diagonal preconditioning, momentum is another key component that contributes to the efficiency of Adam. It is well-known that Nesterov momentum \citep{nesterov1983method} can accelerate the convergence of GD for smooth convex \citep{nesterov2013introductory} and convex-like \citep{hinder2020near} functions up to the rate $\cO(1/T^2)$. Consequently, there is an array of works that aim to establish theoretical guarantees for AdaGrad-type methods with Nesterov acceleration, including the works of \citet{levy2018online,cutkosky2019anytime,kavis2019unixgrad,rodomanov2024universality,kreisler2024accelerated}. However, to the best of our knowledge, all such algorithms achieve accelerated theoretical convergence rates only for scalar stepsizes. Therefore, another natural question appears:
\begin{question}\label{q2}
  Can we design an adaptive preconditioned gradient method that provably benefits from both diagonal AdaGrad-type preconditioning and momentum?
\end{question}
To the best of our knowledge, the only attempt to answer this question was made by \citet{trifonov2025incorporating}. However, they made additional unrealistic assumptions about the dynamics of the preconditioning operator and considered only a smooth and strongly convex, non-stochastic setting. Their theoretical results provided a highly limited explanation of the benefits of preconditioning, including a lack of adaptation to stochasticity and matrix/anisotropic H\"older smoothness.

\subsection{Contributions and Related Work}\label{sec:contrib}

In this paper we give positive answers to \Cref{q1,q2} and provide the following contributions:
\begin{enumerate}[label=\bf(\roman*)]
  \item We develop a unified analysis framework for adaptive preconditioned stochastic gradient methods under the matrix H\"older smoothness and bounded variance. Using this framework, in \Cref{sec:alg}, we provide a single convergence proof that is applicable to most existing AdaGrad-type algorithms, recovering the state-of-the-art convergence guarantees for AdaGrad-Norm, AdaGrad, and ASGO/One-sided Shampoo. Moreover, we establish convergence guarantees for DASGO, a computationally efficient variant of ASGO proposed by \citet{an2025asgo}, and find its fundamental connection with the recently proposed Scion method by \citet{pethick2025training}.

  \item We develop a novel unified analysis of adaptive preconditioned stochastic gradient methods with Nesterov acceleration under the additional assumption that the smoothness and noise operators,\footnote{Refer to \Cref{ass:f,ass:sigma} for precise definitions.} $\mL$ and $\mSigma$, commute with any preconditioner $\mH_k$. In particular, in \Cref{sec:alg2}, we show that the convergence of algorithms with diagonal preconditioning, such as AdaGrad and DASGO, can be significantly improved with no extra assumptions compared to their non-accelerated counterparts. To the best of our knowledge, this is the first theoretical justification that AdaGrad can benefit from both momentum and diagonal preconditioning.
\end{enumerate}

Below we discuss some additional related works.

\textbf{Parameter-free algorithms.}
There is an important research direction aimed at designing parameter-free variants of AdaGrad, which can avoid tuning the parameter $\eta \propto \norm{x^*}$ in \cref{eq:adagrad_norm}. This includes the works of \citet{cutkosky2018black,orabona2021parameter,defazio2023learning,mishchenko2023prodigy,ivgi2023dog,khaled2023dowg,kreisler2024accelerated}.\footnote{Additional references can be found in the overview of \citet{orabona2023normalized}.} However, to the best of our knowledge, the existing results are applicable only to scalar stepsizes, which are rarely used in practice. Designing parameter-free gradient methods with diagonal or matrix preconditioning is an interesting question for future work.

\textbf{Concurrent unified analysis framework.}
\citet{xie2025structured} developed a unified analysis for AdaGrad-type methods, where they also adopt the matrix smoothness assumption. We found their work during the preparation of our literature review, at a point when our results had already been finalized. Although the results of \citet{xie2025structured} share some similarities with ours and are capable of providing a partially positive answer to \Cref{q1}, their analysis has substantial differences and drawbacks. Specifically, it only covers the smooth case and lacks adaptation to non-smooth or H\"older smooth functions. In addition, it requires a more restrictive stochastic gradient noise assumption and, most importantly, does not contain any results about using momentum acceleration, thus completely missing an answer to the fundamental \Cref{q2}.

\textbf{Exponential running average.}
AdaGrad-type algorithms, like RMSProp, often utilize the exponential moving average (EMA): they replace the cumulative sum of the squared gradients $\sum_{i=0}^k\sqn{g_i}$ in \cref{eq:adagrad_norm} with the exponential running average $\sum_{i=0}^k\beta^i\sqn{g_i}$. Notably, EMA is the third and last key component of Adam, in addition to diagonal preconditioning and momentum. Moreover, \citet{defossez2020simple} showed how to analyze AdaGrad with EMA and explained that it is related to the standard AdaGrad in the same way as fixed stepsize SGD is related to decaying stepsize SGD. Consequently, we can develop EMA versions of our algorithms as well as their convergence proofs. However, \citet{defossez2020simple} could not justify the benefits of using momentum. Hence, our theoretical justification of the benefits of momentum and diagonal preconditioning, combined with the results for EMA by \citet{defossez2020simple}, may provide the ultimate explanation for the efficiency of Adam.

\section{Preliminaries}\label{sec:pre}

{\bf Notation.}
$\dim {\cX}$ is the dimension of the space $\cX$;
$\sL$ is the space of linear operators $\cX \to \cX$, for arbitrary operator $\mA \in \sL$, $\mA\adj \in \sL$ denotes its adjoint operator, $\mI \in \sL$ and $\mO \in \sL$ denote the identity and the zero operators, respectively;
$\Sym \subset \sL$ is the space of self-adjoint linear operators, $\Sympp, \Symp \subset \Sym$ are the spaces of positive definite and positive semi-definite self-adjoint operators, respectively; $\prec,\preceq,\succ,\succeq$ denote the standard L\"owner order on $\Sym$;
$\<,>$ and $\norm{}$ denote the standard inner product and Euclidean norm on $\cX$ or $\sL$, depending on the context, in particular, $\<\mA,\mB> = \trace{\mA\mB\adj}$ for $\mA,\mB \in \sL$;
for arbitrary $\mH \in \Sympp$, $\norm{}_\mH$ denotes the weighted Euclidean norm in $\cX$, i.e., $\sqn{x}_\mH = \<x,\mH x>$ for $x \in \cX$;
$\norms{}$ and $\normt{}$ denote the operator and trace norm on $\sL$, respectively, i.e., $\norms{\mA} = \max_{\norm{x}\leq 1}\norm{\mA x}$ and $\normt{\mA} = \trace{\sqrt{\mA\mA\adj}}$ for all $\mA \in \sL$; for arbitrary $y,z \in \cX$, by $z\<y,> \in \sL$ we denote the rank-1 linear operator $x\mapsto \<x,y>z$;
by $\Sym^d \subset \R^{d\times d}$ we denote the space of $d\times d$ symmetric matrices;
by $\odot$, we denote the Hadamard vector or matrix product.

\subsection{Unified Preconditioning Framework}\label{sec:precond}

The preconditioned gradient method uses the update rule in \cref{eq:gd}, which requires the preconditioning operator $\mH_k \in \Sympp$. Similar to the approach of \citet{gupta2017unified}, we restrict the operator $\mH_k$ to belong to a certain subspace of self-adjoint operators $\cH \subset \Sym$. As discussed in \Cref{sec:intro:precond}, we can obtain most existing AdaGrad-type methods by choosing different instances of the space $\cH$. However, to develop a single unified convergence proof for these algorithms, we need to impose formal assumptions on the space $\cH$. This is done through the following \Cref{def:op_f} and \Cref{ass:H}.


\begin{definition}\label{def:op_f}
  Let $\psi(h) \colon I \to \R$ be a scalar function defined on an arbitrary  interval $I \subset \R$. Let $\cS_I \subset \Sym$ be the set of self-adjoint operators, with eigenvalues lying in $I$. The corresponding operator function $\psi(\mH) \colon \cS_I \to \Sym$ is defined as follows:
  \begin{equation}
    \psi(\mH) = \tsum_i \psi(\lambda_i) \mP_i,
  \end{equation}
  where $\mH = \sum_i \lambda_i \mP_i$ is the eigendecomposition of the operator $\mH \in \cS_I$, that is, $\lambda_i \in I$ are the eigenvalues of $\mH$, and $\mP_i \in \Sym$ are the projection operators onto the corresponding eigenspaces.
\end{definition}

\begin{assumption}\label{ass:H}
  The space of linear operators $\cH \subset \Sym$ satisfies the following properties:
  \begin{enumerate}[label=\textnormal{\bf(A\ref*{ass:H}.\arabic*)},ref=\textnormal{A\ref*{ass:H}.\arabic*}]
    \item\label[property]{ass:H:proj} The projection onto $\cH$ is order preserving, that is, $\proj[\cH]{\mH} \in \Sympp$ for all $\mH \in \Sympp$.
    \item\label[property]{ass:H:func} The space $\cH$ is closed under arbitrary operator functions, that is, $\psi(\mH) \in \cH$ for all $\mH \in \cH$ and $\psi(h)\colon \R \to \R$.
  \end{enumerate}
\end{assumption}

Next, according to \citet{gupta2017unified}, we describe a unified way to define the preconditioning operator $\mH_k \in \Sympp$ based on the choice of the space $\cH$. Given the past gradients $g_0,\ldots,g_k \in \cX$, the preconditioning operator $\mH_k$ is defined as a solution to the following optimization problem:
\begin{equation}\label{eq:H}
  \mH_k = \argmin_{\mH \in \cH \cap \Sympp} \<\mH, \mS_k> + \<\mI,\phi(\mH)>,
  \;\;\text{where}\;\; \mS_k = \tsum_{i=0}^{k} g_i\<g_i,>,
\end{equation}
where $\psi(h)\colon \R_{++} \to \R$ is a strictly convex non-negative potential function. The optimization form of this definition allows the use of the standard tool from online optimization, the Follow-the-Leader/Be-the-Leader (FTL-BTL) lemma \citep{kalai2005efficient}. It can be summarized in the following inequality:
\begin{equation}\label{eq:ftl_btl}
  \tsum_{i=-1}^{k} l_i(\theta_i) \leq \tsum_{i=-1}^{k} l_i(\theta_k),
  \;\;\text{where}\;\;
  \theta_i = \targmin_{\theta \in \Theta} l_i(\theta),\tag{FTL-BTL}
\end{equation}
where $l_{-1}(\theta)\clc l_k(\theta)\colon \Theta \to \R$ is an arbitrary sequence of functions defined on a domain $\Theta$.\footnote{The proof of \cref{eq:ftl_btl} can be found in Appendix~A of \citet{gupta2017unified}.}
Similar to \citet{gupta2017unified}, we can use this result to obtain the following \Cref{lem:ftl_btl}, which is one of the key elements in the unified analysis of Adagrad-type algorithms.
\begin{lemma}<lem:ftl_btl>
  The preconditioner $\mH_k$ defined in \cref{eq:H} satisfies the following inequality:
  \begin{equation}
    \tsum_{i=0}^{k} \sqn{g_i}_{\mH_i} \leq \<\mH_k,\mS_k> + \<\mI, \phi(\mH_k)>.
  \end{equation}
\end{lemma}
The application of \Cref{lem:ftl_btl} is not limited to a specific choice of the potential function. However, to obtain Adagrad-type preconditioners, we will use the following potential function $\phi(h)$, which is given as follows:
\begin{equation}\label{eq:phi}
  \phi(h) = \delta \cdot h + \eta^2/h,
\end{equation}
where $\delta,\eta > 0$ are positive parameters. Here appears the first key difference from \citet{gupta2017unified}: using our \Cref{ass:H}, we can explicitly compute the preconditioner $\mH_k$, as stated by the following \Cref{lem:H}.
\begin{lemma}<lem:H>
  The auxiliary problem in \cref{eq:H} with the potential function $\phi(h)$ defined in \cref{eq:phi} has the following unique solution:
  \begin{equation}\label{eq:H_exp}
    \mH_k = \eta \left(\delta \mI + \proj[\cH]{\mS_k}\right)^{-1/2}.
  \end{equation}
  Moreover, the following operator inequality holds:
  \begin{equation}\label{eq:H_order}
    \mH_{k+1} \preceq \mH_{k}.
  \end{equation}
\end{lemma}

Overall, the assumptions that we impose on the space of preconditioning operators $\cH$ (\Cref{ass:H:proj,ass:H:func} in \Cref{ass:H}) are closely related to the notion of a well-structured preconditioner set used by \citet{xie2025structured}. Consequently, the unified analysis of \citet{xie2025structured} shares some similarities with ours but suffers from significant disadvantages discussed in \Cref{sec:contrib}.

\begin{table}[t]
  \centering
  \caption{The linear space $\cX$, the space of preconditioning operators $\cH$ satisfying \Cref{ass:H}, and the (possibly non-Euclidean) norm $\dist{}$ defined in \cref{eq:dist} for AdaGrad-Norm \citep{streeter2010less}, AdaGrad \citep{duchi2011adaptive,mcmahan2010adaptive}, ASGO/One-sided Shampoo \citep{an2025asgo,xie2025structured}, and DASGO \citep{an2025asgo}.}
  \label{tab}
  \begin{NiceTabular}[]{cccc}
    \toprule
    \bf Algorithm &$\cX$& $\cH$ & $\dist{}$
    \\\midrule
    AdaGrad-Norm &$\R^d$& $\set{g \mapsto \beta g \given \beta \in \R} $ & $\frac{1}{\sqrt{d}}\norm{}$
    \\\midrule
    AdaGrad &$\R^d$& $\set{g  \mapsto \bm{b} \odot g \given \bm{b} \in \R^d} $ & $\normi{}$
    \\\midrule
    ASGO/One-sided Shampoo &$\R^{m \times n}$& $\set{G \mapsto B G \given B \in \Sym^m}$ & $\frac{1}{\sqrt{n}}\smax(\argordot{}{})$
    \\\midrule
    DASGO & $\R^{m \times n}$ & $\set{G  \mapsto \diag{\bm{b}} G \given \bm{b} \in \R^m}$ & $\frac{1}{\sqrt{n}}\norm{}_{2\to\infty}$
    \\\bottomrule
  \end{NiceTabular}
\end{table}

\subsection{Assumptions on the Objective Function}\label{sec:ass}

In this section, we formalize the assumptions that we impose on the objective function $f(x)$. The following \Cref{ass:f} formalizes the convexity and matrix H\"older smoothness properties of the function $f(x)$. Note that in the smooth case ($\nu = 1$) \Cref{ass:f} matches the definitions used by \citet{an2025asgo,xie2025structured}. In the non-smooth case ($\nu=0$), is  more general compared to the assumption used by \citet[Corollary~2]{an2025asgo}. Note that \citet{xie2025structured} provides no results in the non-smooth case, and neither of the works of \citet{an2025asgo,xie2025structured} provides results in the H\"older smooth case for $0 < \nu < 1$.

\begin{assumption}\label{ass:f}
  The function $f(x)$ is convex and $(\normt{\mL}^{\frac{1-\nu}{2}},\nu)$-H\"older smooth with respect to the norm $\norm{}_\mL$, where $\nu \in [0,1]$ and $\mL \in \cH \cap \Sympp$. That is, for all $x_1,x_2 \in \cX$ and $\nabla f(x_1) \in \partial f(x_1)$, the following inequalities hold:
  \begin{equation}
    0 \leq f(x_2) - f(x_1) - \<\nabla f(x_1),x_2-x_1> \leq \tfrac{1}{1+\nu}\normt{\mL}^{\frac{1-\nu}{2}}\norm{x_2-x_1}^{1+\nu}_{\mL}.
  \end{equation}
\end{assumption}

Additionally, using the matrix H\"older smoothness property in \Cref{ass:f}, we establish the following \Cref{lem:grad}, which will be further used in our convergence analysis.

\begin{lemma}<lem:grad>
  For all $x \in \cX$ and $\nabla f(x) \in \partial f(x)$, the following inequality holds:
  \begin{equation}
    \norm{\nabla f(x)}_{\mL^{-1}}^{2}
    \leq
    \left(\tfrac{1+\nu}{\nu}\right)^{\frac{2\nu}{1+\nu}}
    \normt{\mL}^{\frac{1-\nu}{1+\nu}}
    \left(f(x) - f(x^*)\right)^{\frac{2\nu}{1+\nu}},
  \end{equation}
  where in the case $\nu = 0$, we use the convention $0^0 = 1$.
\end{lemma}

The matrix smoothness in \Cref{ass:f} is also closely related to the non-Euclidean smoothness property, which recently received a lot of attention \citep{bernstein2024old,pethick2025training,kovalev2025understanding,riabinin2025gluon} due to the practical success of the Muon optimizer \citep{jordan2024muon}. Let function $\dist{x}\colon\cX \to \R_+$ be defined as follows:
\begin{equation}\label{eq:dist}
  \dist{x} = \norms{\proj[\cH]{\mX}}^{1/2},\;\;\;\text{where}\;\;\mX=x\<x,>.
\end{equation}
One can verify that the function $\dist{x}$ is a norm on the linear space $\cX$, as shown in \Cref{lem:dist}. Besides, \Cref{ass:f} implies that the function $f(x)$ is $(\normt{L},\nu)$-H\"older smooth with respect to this possibly non-Euclidean norm $\dist{}$. That is, the following inequality holds for all $x_1,x_2 \in \cX$:
\begin{equation}\label{eq:Holder}
  f(x_2) - f(x_1) - \<\nabla f(x_1),x_2-x_1> \leq \tfrac{1}{1+\nu}\normt{\mL}\left[\dist{x_2-x_1}\right]^{1+\nu}.
\end{equation}
We provide additional discussion of the connection between \Cref{ass:f} and the non-Euclidean H\"older smoothness in \cref{eq:Holder} in \Cref{sec:alg}.
\begin{lemma}<lem:dist>
  The function $\dist{x}$ defined in \cref{eq:dist} is a norm. That is, it is subadditive, absolutely homogeneous, non-negative, and positive definite.
\end{lemma}

Additionally, we provide the assumptions on the stochastic gradient noise in the following \Cref{ass:sigma}. These are more general than the assumptions used by both \citet{an2025asgo} and \citet{xie2025structured}. In particular, they assume the ordering $\E[\xi \sim \cD]{n(x;\xi)\<n(x;\xi),>} \preceq \mSigma^2$, which implies \Cref{ass:sigma:variance}, and hence, is more restrictive.

\begin{assumption}\label{ass:sigma}
  There exists a stochastic estimator $\nabla f(x;\xi) = n(x;\xi) + \nabla f(x)$ of the (sub)gradient $\nabla f(x) \in \partial f(x)$ of the objective function $f(x)$, where $n(x;\xi)$ is the noise and $\xi \sim \cD$ is a random variable. The noise $n(x;\xi)$ satisfies the following properties:
  \begin{enumerate}[label=\textnormal{\bf(A\ref*{ass:sigma}.\arabic*)},ref=\textnormal{A\ref*{ass:sigma}.\arabic*}]
    \item\label[property]{ass:sigma:mean} Zero mean: $\E[\xi \sim \cD]{n(x;\xi)} = 0$ for all $x \in \cX$.
    \item\label[property]{ass:sigma:variance} Bounded variance: $\E[\xi \sim \cD]{\sqn{n(x;\xi)}_{\mSigma^{-1}}} \leq \normt{\mSigma}$\; for all $x \in \cX$, where $\mSigma \in \cH \cap \Sympp$.
  \end{enumerate}
\end{assumption}

\section{Unified Analysis of Adaptive SGD with Preconditioning}\label{sec:alg}

\subsection{General Algorithm and its Convergence}

\begin{algorithm}[t]
  \caption{Adaptive SGD with Preconditioning}
  \label{alg}
  \begin{algorithmic}[1]
    \State\textbf{input:} $x_0 \in \cX$, $K \in \set{1,2,\ldots}$
    \For{$k=0\clc K$}
    \State sample $\xi_k \sim \cD$
    \State compute $g_k = \nabla f(x_k;\xi_k)$
    \State compute $\mH_k \in \cH \cap \Sympp$ using \cref{eq:H,eq:H_exp}
    \State compute $x_{k+1} \in \cX$ using \cref{eq:gd}.
    \EndFor
    \State\textbf{output:} $\ox_K=\tfrac{1}{K+1}\sum_{k=0}^{K}x_k$\label{line:ox}
  \end{algorithmic}
\end{algorithm}

Based on the discussion in \Cref{sec:precond}, we formalize the adaptive stochastic gradient method with preconditioning as \Cref{alg}. In this section, we develop the unified convergence analysis of this algorithm. First, we obtain an upper bound on the expected regret $\E{\tsum_{k=0}^Kf(x_k) - f(x^*)}$ in the following \Cref{lem:f_bound}. The proof of this lemma, in many ways, relies on the previously obtained \Cref{lem:ftl_btl,lem:H}.

\begin{lemma}<lem:f_bound>
  Under the conditions of \Cref{thm:alg}, the following inequality holds:
  \begin{equation}
    \tsum_{k=0}^K\E{f(x_k) - f(x^*)}
    \leq
    \tfrac{3}{2}\cR\<\mI,\proj[\cH]{\E{\mS_{K}}}^{1/2}>
    +\tfrac{3}{2}\sqrt{\delta}\cR\dim{\cX}.
  \end{equation}
\end{lemma}

Next, in the following \Cref{lem:S_bound}, we establish an upper bound on the right-hand side of the inequality in \Cref{lem:f_bound}, using \Cref{ass:sigma} and the previously obtained \Cref{lem:grad}.

\begin{lemma}<lem:S_bound>
  Under the conditions of \Cref{thm:alg}, the following inequality holds:
  \begin{equation}
    \begin{aligned}
      \<\mI,\proj[\cH]{\E{\mS_K}}^{1/2}>
      &\leq
      \sqrt{K+1}^{\tfrac{1-\nu}{1+\nu}}\normt{\mL}^{\frac{1}{1+\nu}}
      \left[\tsum_{k=0}^K\E{f(x_k) - f(x^*)}\right]^{\frac{\nu}{1+\nu}}
      \\&
      +\sqrt{K+1}\normt{\mSigma}.
    \end{aligned}
  \end{equation}
\end{lemma}

Finally, with the help of \Cref{lem:f_bound,lem:S_bound}, we obtain the convergence result for \Cref{alg} in the following \Cref{thm:alg}. Note that this result requires the inequality in \cref{eq:R} to hold almost surely, which may not be satisfied, especially in the stochastic setting. However, this issue can be easily resolved with an additional projection step at each iteration. Refer to \Cref{sec:clip} for details.

\begin{theorem}<thm:alg>
  Under \Cref{ass:H,,ass:f,,ass:sigma}, let $\eta = \cR$, where $\cR > 0$ almost surely satisfies the following inequality:
  \begin{equation}\label{eq:R}
    \max_{k=0,\ldots,K}\dist{x_k - x^*} \leq \cR.
  \end{equation}
  Then, the output $\ox_K \in \cX$ of \Cref{alg} satisfies the following inequality:
  \begin{equation}
    \E{f(\ox_K) - f(x^*)}
    \leq
    \frac{3\normt{\mL}\cR^{1+\nu}}{(K+1)^{\frac{1+\nu}{2}}}
    +\frac{3\normt{\mSigma}\cR}{\sqrt{K+1}}
    +\frac{3\sqrt{\delta}\cR\dim{\cX}}{(K+1)}.
  \end{equation}
\end{theorem}

\subsection{Related Algorithms}\label{sec:cases}

In this section, we discuss the connection of \Cref{alg} with existing adaptive gradient methods with preconditioning.

\textbf{Connection with AdaGrad-Norm, AdaGrad, and ASGO/One-sided Shampoo.}
We can obtain AdaGrad-Norm, AdaGrad, ASGO/One-sided Shampoo as special instances of \Cref{alg} by choosing the space of preconditioning operators $\cH$ satisfying \Cref{ass:H} according to \Cref{tab}. In the case $\nu=1$, \Cref{thm:alg} recovers the state-of-the-art convergence guarantees for AdaGrad under anisotropic smoothness \citep{liu2024adagrad} and for ASGO/One-sided Shampoo \citep{an2025asgo,xie2025structured} under matrix smoothness. However, recall that \citet{liu2024adagrad,an2025asgo,xie2025structured} require a more restrictive noise variance bound as discussed in \Cref{sec:ass}, and do not cover H\"older smoothness. In contrast, \Cref{thm:alg} works for arbitrary $\nu \in [0,1]$, which implies that \Cref{alg} can adapt to different levels of anisotropic/matrix smoothness.

\textbf{Connection with DASGO.}
Notably, \Cref{alg} recovers DASGO, a lightweight version of ASGO/One-sided Shampoo that uses diagonal preconditioning and was proposed by \citet{an2025asgo} without any convergence guarantees. Consequently, \Cref{thm:alg} provides the first convergence guarantees for DASGO, to the best of our knowledge. Moreover, in \Cref{sec:alg2}, we will show that the convergence of DASGO, as well as AdaGrad, can be accelerated using Nesterov momentum.

\textbf{Connection between ASGO/One-sided Shampoo and Muon.}
Recently, \citet{jordan2024muon} proposed using the Shampoo optimizer \citep{gupta2018shampoo} with gradient accumulation turned off. This led to the development of Muon, a new optimizer with promising practical performance. The convergence of Muon was analyzed from the perspective of gradient methods with the non-Euclidean matrix spectral norm by \citet{bernstein2024old,pethick2025training,kovalev2025understanding}. Notably, our analysis captures the connection between ASGO/One-sided Shampoo and non-Euclidean optimization with the spectral norm. Indeed, as discussed in \Cref{sec:ass}, \Cref{ass:f} implies the $(\normt{\mL},\nu)$-H\"older smoothness in \cref{eq:Holder} with respect to the norm $\dist{}$, which, according to \Cref{tab}, coincides with the matrix spectral norm (up to constant factors). Moreover, in the case of ASGO/One-sided Shampoo, \Cref{thm:alg} provides the convergence result in terms of the constant $\normt{\mL}$ and the norm $\dist{}$.

\textbf{Connection between DASGO and Scion.}
Recently, \citet{pethick2025training} proposed Scion, a new variant of Muon, which, instead of the spectral norm, can use the matrix norm $\norm{}_{2\to\infty}$: the maximal Euclidean norm of a row of a matrix. Note that in the case of DASGO, the norm $\dist{}$ defined in \cref{eq:dist} coincides with the norm $\norm{}_{2\to\infty}$ up to multiplicative constants, according to \Cref{tab}. Hence, Scion with the norm $\norm{}_{2\to\infty}$ can be obtained by turning off the gradient accumulation in DASGO, that is, choosing $\mS_k = g_k\<g_k,>$ in \cref{eq:H}. In other words, DASGO is connected to Scion in the same way as ASGO/(One-sided) Shampoo is connected to Muon. It is important to highlight that the iterations of Shampoo are not cheap and require matrix inversions, which triggered the development of the computationally effective alternative, Muon, by \citet{jordan2024muon}. However, the iterations of DASGO are not only inexpensive, but they also utilize adaptive preconditioning and have much more attractive theoretical convergence properties compared to Scion. Hence, it is worth trying to use DASGO in the practical scenarios identified by \citet{pethick2025training} to benefit from using the non-Euclidean norm $\norm{}_{2\to\infty}$.

\section{SGD with Preconditioning and Acceleration}\label{sec:alg2}

\subsection{General Accelerated Algorithm and its Convergence}

\begin{algorithm}[t]
  \caption{Accelerated Adaptive SGD with Preconditioning}
  \label{alg2}
  \begin{algorithmic}[1]
    \State\textbf{input:} $x_0=\ox_0 \in \cX$, $K \in \set{1,2,\ldots}$
    \For{$k=0\clc K$}
    \State sample $\xi_k \sim \cD$
    \State compute $g_k = \nabla f_k(x_k;\xi_k)$, where $f_k(x)$ is defined in \cref{eq:f_k}
    \State compute $\mH_k \in \cH \cap \Sympp$ using \cref{eq:H,eq:H_exp}
    \State compute $x_{k+1} \in \cX$ using \cref{eq:gd}.
    \State compute $\ox_{k+1} = \alpha_k x_{k+1} + (1-\alpha_k)\ox_k$\label{line2:ox}
    \EndFor
    \State\textbf{output:} $\ox_{K+1}$
  \end{algorithmic}
\end{algorithm}

In this section, we develop accelerated adaptive SGD with preconditioning, which is summarized in \Cref{alg2}, and provide its unified convergence analysis. First, to simplify the analysis, we use the interpretation of Nesterov momentum acceleration \citep{nesterov1983method} by \citet{kovalev2024linear}. The idea is that we define the functions $f_k(x)\colon \cX \to \R$ as follows:
\begin{equation}\label{eq:f_k}
  f_k(x) =\alpha_k^{-2} \cdot f(\alpha_k x + (1-\alpha_k)\ox_k),
  \quad\text{where}\;\;
  \alpha_k \in (0,1]
  \text{\;\;and\;\;}
  \ox_k \in \cX,
\end{equation}
where $\ox_k \in \cX$ is updated according to \cref{line2:ox} at each iteration. We then apply the preconditioned SGD iterations in \cref{eq:gd} to this ``time-varying'' function $f_k(x)$. With this approach, we can upper-bound the expected objective function suboptimality $\E{f(\ox_{K+1}) - f(x^*)}$ using the expected regret-like sum $\tsum_{k=0}^K \E{f_k(x_{k+1}) - f_k(x^*)}$ in the following \Cref{lem:alpha}.

\begin{lemma}<lem:alpha>
  Under the conditions of \Cref{thm:alg2}, the following inequality holds:
  \begin{equation}
    \tfrac{1}{4}(K+2)^2\E{f(\ox_{K+1}) - f(x^*)} \leq \tsum_{k=0}^K \E{f_k(x_{k+1}) - f_k(x^*)}.
  \end{equation}
\end{lemma}

Next, we proceed with the additional \Cref{ass:LH} on the operators $\mL,\mSigma \in \cH$ defined in \Cref{ass:f,ass:sigma}. It is important to highlight that this assumption always holds when the space of preconditioners $\cH$ contains only diagonal operators. Hence, this assumption is automatically satisfied for algorithms with diagonal preconditioning like AdaGrad and DASGO.

\begin{assumption}\label{ass:LH}
  The operators $\mL \in \cH$ in \Cref{ass:f} and $\mSigma \in \cH$ in \Cref{ass:sigma} commute with the space $\cH$, that is, $\mL\mH = \mH\mL$ and $\mSigma\mH = \mH\mSigma$ for all $\mH \in \cH$.
\end{assumption}

The key idea for the analysis of \Cref{alg2} is that under \Cref{ass:LH}, the square of the precondition operator $\mH_k$, defined in \cref{eq:H_exp}, is a solution to the optimization problem in \cref{eq:LH}, as indicated by \Cref{lem:LH}. Hence, similar to the analysis of the non-accelerated \Cref{alg}, we can utilize the FTL-BTL lemma \eqref{eq:ftl_btl} and obtain one of the key inequalities in \Cref{lem:ftl_btl2}.

\begin{lemma}<lem:LH>
  Under \Cref{ass:LH}, the operator $\mH_k^2$ defined by \cref{eq:H_exp} is a solution to the following problem, where $\mB = \mL$ or $\mB = \mSigma$:
  \begin{equation}\label{eq:LH}
    \mH_k^2 \in \argmin_{\mQ \in \cH \cap \Sympp} \<\mQ,\mB\mS_k> + \<\mB,\delta\mQ - \eta^2 \ln(\mQ)>.
  \end{equation}
\end{lemma}
\begin{lemma}<lem:ftl_btl2>
  Under \Cref{ass:LH}, the following inequality holds for $\mB = \mL$ or $\mB = \mSigma$:
  \begin{equation}
    \E*{\tsum_{i=0}^{k} \sqn{g_i}_{\mB\mH_i^2}} \leq
    \eta^2\normt{\mB}\ln\left[\tfrac{1}{\delta}\eta^2\left(\E{\normt{\mH_k^{-1}}}\right)^2\right].
  \end{equation}
\end{lemma}

Finally, using the inequality in \Cref{lem:ftl_btl2}, we obtain the key upper bound on the regret-like sum $\E{f_k(x_{k+1}) - f_k(x^*)}$ in \Cref{lem:f_bound2}.

\begin{lemma}<lem:f_bound2>
  Under the conditions of \Cref{thm:alg2}, the following inequality holds:
  \begin{equation}
    \begin{aligned}
      \tsum_{k=0}^K \E{f_k(x_{k+1}) - f_k(x^*)} &\leq
      \tfrac{1}{4}\cC_K(K+2)^{\frac{3(1-\nu)}{2}}\normt{\mL}
      \cR^{1+\nu}
      \\&
      +\tfrac{1}{4}\cC_K(K+2)^{\frac{3}{2}}\normt{\mSigma}
      \cR
      +\sqrt{\delta}\cR\dim{\cX}.
    \end{aligned}
  \end{equation}
\end{lemma}

Now, all that remains is to combine \Cref{lem:f_bound2} with \Cref{lem:alpha} and obtain the main convergence result for \Cref{alg2} in \Cref{thm:alg2}. Similar to the non-accelerated result in \Cref{thm:alg}, we require the inequality in \cref{eq:R} to hold almost surely. This can be easily guaranteed by an additional projection step at each iteration, as discussed in \Cref{sec:clip}.

\begin{theorem}<thm:alg2>
  Under \Cref{ass:H,,ass:f,,ass:sigma,,ass:LH}, let $\eta = 2\cR$, where $\cR > 0$ satisfies \cref{eq:R}, and let $\alpha_k = 2/(k+2)$.
  Then, the output $\ox_{K+1} \in \cX$ of \Cref{alg2} satisfies the following inequality:
  \begin{equation}
    \E{f(\ox_{K+1}) - f(x^*)}
    \leq
    \frac{\cC_K\normt{\mL}\cR^{1+\nu}}{(K+2)^{\frac{1+3\nu}{2}}}
    +\frac{\cC_K\normt{\mSigma}\cR}{\sqrt{K+2}}
    +\frac{4\sqrt{\delta}\cR\dim{\cX}}{(K+2)^2},
  \end{equation}
  where the constant $\cC_K > 0$ satisfies the following relation:
  \begin{equation}\label{eq:C}
    \cC_K = \cO\left(
      1
      + \ln K
      + \ln \tfrac{\normt{\mL}\cR^\nu}{\sqrt{\delta}}
      + \ln \tfrac{\normt{\mSigma}}{\sqrt{\delta}}
    \right).
  \end{equation}
\end{theorem}

\subsection{AdaGrad and DASGO with Momentum Acceleration}

In this section, we provide a detailed discussion of our results for two special instances of adaptive gradient methods with diagonal preconditioning: AdaGrad and DASGO. In the case of DASGO, let $\cX = \R^{m\times n}$ be the space of $m\times n$ matrices and consider the following special instance of \cref{eq:main}:
\begin{equation}
  \min_{X \in \R^{m\times n}} f(X).
\end{equation}
We choose the space $\cH$ of preconditioning operators $\R^{m\times n} \to \R^{m\times n}$ for DASGO according to \Cref{tab}. That is, $\cH = \set{G \mapsto \diag{\bm{b}} G \given \bm{b} \in \R^m}$, which obviously satisfies \Cref{ass:H}. Note that AdaGrad can be obtained from DASGO by simply choosing $n=1$. Henceforth, for simplicity, we will consider only DASGO.

Next, we specialize \Cref{ass:f,ass:sigma} to the setting of DASGO. In particular, we define the operator $\mL\in \cH$ in \Cref{ass:f} as $\mL\colon X \mapsto n^{\frac{\nu-1}{2}}\diag{\bm{l}}X$, where $\bm{l} = (\bm{l}_1,\ldots,\bm{l}_m) \in \R^m_{++}$ and $X \in \R^{m\times n}$. For example, in the case $\nu=1$ and $n=1$, \Cref{ass:f} exactly matches the anisotropic smoothness assumption used by \citet{liu2024adagrad}. In the general case $\nu \in [0,1]$ and $n \geq 1$, \Cref{ass:f} implies the $(\norm{\bm{l}}_1,\nu)$-H\"older smoothness with respect to the non-Euclidean norm $\norm{}_{2\to\infty}$, that is, the following special instance of the inequality in \cref{eq:Holder} holds:
\begin{equation}
  0 \leq f(X_2) - f(X_1) - \<\nabla f(X_1),X_2-X_1> \leq \tfrac{1}{1+\nu}
  \norm{\bm{l}}_1^{\frac{1-\nu}{2}}\norm{X_2-X_1}^{1+\nu}_{2\to\infty}.
\end{equation}
Similarly, we define the operator $\mSigma \in \cH$ in \Cref{ass:sigma:variance} as $\mSigma\colon X \mapsto n^{-\frac{1}{2}}\diag{\bm{\sigma}}X$, where $\bm{\sigma} = (\bm{\sigma}_1,\ldots,\bm{\sigma}_m) \in \R^m_{++}$ and $X \in \R^{m\times n}$. Consequently, the variance bound in \Cref{ass:sigma:variance} turns into the following inequality:
\begin{equation}
  \E[\xi \sim \cD]{\tsum_{i=1}^m (1/\bm{\sigma}_i)\sqn{N_i}} \leq \norm{\bm{\sigma}}_1,
  \;\;\text{where}\;\;
  [N_1,\ldots,N_m]^\top = \nabla F(X;\xi) - \nabla F(X).
\end{equation}
This inequality is implied, for instance, by the anisotropic noise assumption used by \citet{liu2024adagrad}, and hence, it is more general.

Further, for simplicity in the presentation of the results, we use the convergence guarantees from \Cref{sec:clip} for the algorithms with projection steps.
Using \Cref{thm:algp} and assuming $\delta \ll 1$, we obtain the following convergence guarantees for AdaGrad and DASGO:
\begin{equation}
  \E{f(\overline{X}_K) - f(X^*)} \leq \tilde\cO\left(\frac{\norm{\bm{l}}_1\norm{X^*}_{2\to\infty}^{1+\nu}}{K^{\frac{1+\nu}{2}}} + \frac{\norm{\bm{\sigma}}_1\norm{X^*}_{2\to\infty}}{\sqrt{K+1}}\right).
\end{equation}
This matches the result of \citet{liu2024adagrad} for AdaGrad in the smooth case ($\nu=1$ and $n=1$), but also provides convergence guarantees for DASGO.
Similarly, using \Cref{thm:algp2}, we establish convergence guarantees for AdaGrad and DASGO with Nesterov momentum:
\begin{equation}\label{eq:accelerated_dasgo}
  \E{f(\overline{X}_{K+1}) - f(X^*)} \leq \tilde\cO\left(\frac{\norm{\bm{l}}_1\norm{X^*}_{2\to\infty}^{1+\nu}}{K^{\frac{1+3\nu}{2}}} + \frac{\norm{\bm{\sigma}}_1\norm{X^*}_{2\to\infty}}{\sqrt{K+1}}\right),
\end{equation}
which substantially improves upon the non-accelerated result above. We can also compare this result with the state-of-the-art result of \citet{kavis2019unixgrad,rodomanov2024universality} for scalar AdaGrad-type stepsizes under the above assumptions:
\begin{equation}\label{eq:accelerated_adagrad_norm}
  \E{f(\overline{X}_{K+1}) - f(X^*)} \leq \tilde\cO\left(\frac{\normi{\bm{l}}\norm{X^*}^{1+\nu}}{K^{\frac{1+3\nu}{2}}} + \frac{\sqrt{m}\normi{\bm{\sigma}}\norm{X^*}}{\sqrt{K+1}}\right).
\end{equation}
Our result in \cref{eq:accelerated_dasgo} is substantially better than the existing result in \cref{eq:accelerated_adagrad_norm} as long as $\norm{\bm{l}}_1 \sim \normi{\bm{l}}$, $\norm{\bm{\sigma}}_1 \sim \normi{\bm{\sigma}}$, and $\norm{X^*} \gg \norm{X^*}_{2\to\infty}$. For instance, in the AdaGrad case ($n=1$), this holds when $\bm{l}$ and $\bm{\sigma}$ are sparse and $X^*$ is dense, which aligns with the conclusions made by \citet{liu2024adagrad} for AdaGrad without momentum acceleration.

\newpage

\cite*{}

\bibliographystyle{apalike}
\bibliography{references}

\newpage

\appendix

\part*{Appendix}

\section{Motivation for Convex Setting}\label{sec:convex}

In this paper, we focus on the case where the objective function $f(x)$ in \cref{eq:main} is convex. There are multiple reasons for this assumption. First, optimization algorithms for convex functions hold substantial practical interest because empirical studies \citep{zhou2019sgd,kleinberg2018alternative} suggest that deep neural networks may adhere to convexity or its variants. Second, for general non-convex functions, it is impossible to achieve meaningful global convergence beyond vague first-order stationarity \citep{carmon2020lower}. However, in practice, it is typically desirable to achieve small values of the objective function, which can only be guaranteed under additional assumptions, such as gradient domination \citep{fatkhullin2022sharp}, star/quasar convexity \citep{hinder2020near}, etc. Such assumptions are, in turn, relaxations of the convexity property itself. Hence, it is natural to consider the convex setting first before trying to relax it. Finally, convex optimization serves as a large source of inspiration for designing efficient optimization algorithms. Notably, many optimization techniques that have practical benefits were initially theoretically justified for convex functions. These include momentum acceleration \citep{nesterov2013introductory}, local training \citep{mishchenko2022proxskip}, and AdaGrad \citep{duchi2011adaptive}, on which Adam itself is based.

\section{Algorithms with Weight Clipping}\label{sec:clip}

\newcommand{\fix}[1]{{\color{red}#1}}

\begin{algorithm}[t]
  \caption{Adaptive SGD with Preconditioning \fix{and Weight Clipping}}
  \label{algp}
  \begin{algorithmic}[1]
    \State\textbf{input:} $x_0 \in \fix{\cQ_\cR}$, $K \in \set{1,2,\ldots}$\label{linep:input}
    \For{$k=0\clc K$}
    \State sample $\xi_k \sim \cD$
    \State compute $g_k = \nabla f(x_k;\xi_k)$
    \State compute $\mH_k \in \cH \cap \Sympp$ using \cref{eq:H,eq:H_exp}
    \State compute $x_{k+1} \in \cX$ using \fix{\cref{eq:gdp}}\label{linep:gd}
    \EndFor
    \State\textbf{output:} $\ox_K=\tfrac{1}{K+1}\sum_{k=0}^{K}x_k$
  \end{algorithmic}
\end{algorithm}

\begin{algorithm}[t]
  \caption{Accelerated Adaptive SGD with Preconditioning \fix{and Weight Clipping}}
  \label{algp2}
  \begin{algorithmic}[1]
    \State\textbf{input:} $x_0=\ox_0 \in \fix{\cQ_\cR}$, $K \in \set{1,2,\ldots}$\label{linep2:input}
    \For{$k=0\clc K$}
    \State sample $\xi_k \sim \cD$
    \State compute $g_k = \nabla f_k(x_k;\xi_k)$, where $f_k(x)$ is defined in \cref{eq:f_k}
    \State compute $\mH_k \in \cH \cap \Sympp$ using \cref{eq:H,eq:H_exp}
    \State compute $x_{k+1} \in \cX$ \fix{and $x_{k+1/2} \in \cX$} using \fix{\cref{eq:gdp}}\label{linep2:gd}
    \State compute $\ox_{k+1} = \alpha_k x_{\fix{k+1/2}} + (1-\alpha_k)\ox_k$\label{linep2:ox}
    \EndFor
    \State\textbf{output:} $\ox_{K+1}$
  \end{algorithmic}
\end{algorithm}

The upper bounds on the expected functional suboptimality in \Cref{thm:alg} for \Cref{alg} and in \Cref{thm:alg2} for \Cref{alg2} require the inequality in \cref{eq:R} to hold almost surely. However, this requirement may not be satisfied, for instance, in the stochastic case. It is important to higlight that such issue is not an artifact of our analysis but a common phenomenon in AdaGrad-type algorithms \citep{duchi2011adaptive,gupta2018shampoo,liu2024adagrad,an2025asgo,xie2025structured}. To bypass this issue, a typical approach is to modify the preconditioned gradient update rule in \cref{eq:gd} by adding an extra projection step onto the set $\cQ_\cR = \set{x \in \cX \given \dist{x} \leq \cR}$, where $\cR > \dist{x^*}$. The modified update rule is given as follows:
\begin{equation}\label{eq:gdp}
  x_{k+1} = \argmin_{x\in\cQ_\cR} \tfrac{1}{2}\sqn{x-x_{k+1/2}}_{\mH_k^{-1}},
  \quad
  x_{k+1/2} = \argmin_{x \in \cX} \<g_k,x> + \tfrac{1}{2}\sqn{x - x_k}_{\mH_k^{-1}}.
\end{equation}
Note that the set $\cQ_\cR$ is convex and hence the projection step is well-defined. Also, note that the projection is performed with respect to the weighted Euclidean norm $\norm{}_{\mH_k^{-1}}$, which may be expensive, for instance, when the preconditioner $\mH_k$ is dense. However, this projection step can be computed efficiently when the preconditioner $\mH_k$ is diagonal. For instance, the projection is equivalent to the coordinate-wise clipping $t \mapsto \min\{\cR,\max\{-\cR,t\}\}$ in AdaGrad, and to the row-wise or column-wise norm-clipping $z \mapsto \min\{1,\cR/\norm{z}\}z$ in DASGO. Below, we discuss the modified update rule \cref{eq:gdp} in relation to the non-accelerated \Cref{alg} and the accelerated \Cref{alg2} in detail, including the additional modifications in \Cref{alg,alg2} and the modifications in the convergence proofs.

{\bf Non-accelerated \Cref{alg} $\to$ \Cref{algp}.} The only modifications to \Cref{alg} are the initialization $x_0 \in \cQ$ on \cref{linep:input} and the modified update rule~\eqref{eq:gdp} on \cref{linep:gd} in \Cref{algp}, as discussed above.
We also modify the proof of \Cref{lem:f_bound} in \Cref{proof:lem:f_bound} by obtaining the following:
\begin{equation}\label{eq:r_kp}
  \tfrac{1}{2}\sqn{x_{k+1} - x^*}_{\mH_k^{-1}}
  \at{uses the update rule for $x_{k+1}$ in \cref{eq:gdp}, the non-expansiveness of the projection, and the fact that $x^* \in \cQ_\cR$}\leq
  \tfrac{1}{2}\sqn{x_{k+1/2} - x^*}_{\mH_k^{-1}}
  \at{uses the update rule for $x_{k+1/2}$ in \cref{eq:gdp}}=
  \tfrac{1}{2}\sqn{x_{k} - x^*}_{\mH_k^{-1}}
  -\<g_k,x_k-x^*> + \tfrac{1}{2}\sqn{g_k}_{\mH_k},
\end{equation}
where \annotate.
One can observe that this \cref{eq:r_kp} coincides with \cref{eq:r_k} in \Cref{proof:lem:f_bound}. Moreover, the inequality \cref{eq:R} holds almost surely due to the projection step in \cref{eq:gdp}. Therefore, the rest of the proof of \Cref{thm:alg} remains unchanged, and we obtain the following \Cref{thm:algp}.
\begin{theorem}<thm:algp>
  Under \Cref{ass:H,,ass:f,,ass:sigma}, let $\eta = \cR$, where $\cR > \dist{x^*}$.
  Then, the output $\ox_K \in \cX$ of \Cref{algp} satisfies the following inequality:
  \begin{equation}
    \E{f(\ox_K) - f(x^*)}
    \leq
    \frac{3\normt{\mL}\cR^{1+\nu}}{(K+1)^{\frac{1+\nu}{2}}}
    +\frac{3\normt{\mSigma}\cR}{\sqrt{K+1}}
    +\frac{3\sqrt{\delta}\cR\dim{\cX}}{(K+1)}.
  \end{equation}
\end{theorem}

{\bf Accelerated \Cref{alg2} $\to$ \Cref{algp2}.} Similarly, to the non-accelerated algorithm, the accelerated \Cref{algp2} contains the modified initialization $x_0 = \ox_0 \in \cQ_\cR$ on \cref{linep2:input} and the modified update rule~\eqref{eq:gdp} on \cref{linep2:gd} in \Cref{algp2}. In addition to the modified \cref{eq:r_kp}, we also modify the first inequality in the proof of \Cref{lem:f_bound2} in \Cref{proof:lem:f_bound2} as follows:
\begin{equation}\label{eq:descentp}
  \E{f_k(x_{\fix{k+1/2}})}
  \leq
  \E*{f_k(x_k) - \sqn{g_k}_{\mH_k} + \<n_k,\mH_k g_k> + \tfrac{1}{1+\nu}\alpha_k^{\nu-1}\normt{\mL}^{\frac{1-\nu}{2}}\norm{g_k}_{\mL\mH_k^2}^{1+\nu}}.
\end{equation}
Here, the only difference is the left-hand side $\E{f_k(x_{k+1/2})}$ compared to $\E{f_k(x_{k+1})}$ in \Cref{proof:lem:f_bound2}, which means that we also have to modify the update rule for $\ox_{k+1}$ on \cref{linep2:ox} of \Cref{algp2} and apply trivial changes to \Cref{lem:alpha}. The rest of the proof of \Cref{thm:alg2} remains unchanged and we obtain the following \Cref{thm:algp2}.
\begin{theorem}<thm:algp2>
  Under \Cref{ass:H,,ass:f,,ass:sigma,,ass:LH}, let $\eta = 2\cR$, where $\cR > \dist{x^*}$, and let $\alpha_k = 2/(k+2)$.
  Then, the output $\ox_{K+1} \in \cX$ of \Cref{algp2} satisfies the following inequality:
  \begin{equation}
    \E{f(\ox_{K+1}) - f(x^*)}
    \leq
    \frac{\cC_K\normt{\mL}\cR^{1+\nu}}{(K+2)^{\frac{1+3\nu}{2}}}
    +\frac{\cC_K\normt{\mSigma}\cR}{\sqrt{K+2}}
    +\frac{4\sqrt{\delta}\cR\dim{\cX}}{(K+2)^2},
  \end{equation}
  where the constant $\cC_K > 0$ satisfies the following relation:
  \begin{equation}
    \cC_K = \cO\left(
      1
      + \ln K
      + \ln \tfrac{\normt{\mL}\cR^\nu}{\sqrt{\delta}}
      + \ln \tfrac{\normt{\mSigma}}{\sqrt{\delta}}
    \right).
  \end{equation}
\end{theorem}

\newpage

\section{Proofs for \Cref{sec:pre}}

\proofsubsection{lem:ftl_btl}

Let functions $l_{-1}(\mH)\clc l_k(\mH)\colon \Sympp \to \R$ be defined as follows:
\begin{equation}
  l_{-1}(\mH) = \<\mI,\phi(\mH)>,\quad
  l_i(\mH) = \sqn{g_i}_\mH \text{\;\;for\;\;}i=0\clc k.
\end{equation}
Let $\mH_{-1}\in\cH\cap \Sympp$ be defined as follows:
\begin{equation}
  \mH_{-1} = \argmin_{\mH \in \cH\cap \Sympp} \<\mI,\phi(\mH)>.
\end{equation}
From \cref{eq:H}, it is easy to verify that the following relation holds for all $i = -1\clc k$:
\begin{equation*}
  \mH_i = \argmin_{\mH \in \cH \cap \Sympp} \tsum_{i=-1}^{k} l_i(\mH).
\end{equation*}
Next, we get the following inequality:
\begin{equation*}
  \tsum_{i=0}^{k} l_i(\mH_i)
  \at{uses the assumption that the potential function $\phi(h)$ is non-negative}\leq
  \tsum_{i=-1}^{k} l_i(\mH_i)
  \at{uses \cref{eq:ftl_btl}}\leq
  \tsum_{i=-1}^{k} l_i(\mH_k).
\end{equation*}
where \annotate.
It remains to to do rearranging and use the definition of the functions $l_i(\mH)$.
\qed

\proofsubsection{lem:H}

First, using \Cref{ass:H:proj,ass:H:func}, we can show that $\mH_k \in \cH\cap\Sympp$.
Next, we show that $\mH_k$ in \cref{eq:H_exp} is a solution to the problem in \cref{eq:H} by verifying the first-order optimality condition:
\begin{align*}
  \nabla(\<, \mS_k> + \<\mI,\phi(\cdot)>)(\mH_k)
  &\at{uses the standard operator function calculus \citep{carlen2010trace}}=
  \mS_k + \phi'(\mH_k)
  \\&\at{uses \cref{eq:phi}}=
  \mS_k + \delta \mI - \eta^2 \mH_k^{-2}
  \\&\at{uses \cref{eq:H_exp}}=
  \mS_k -\proj[\cH]{\mS_k}
  \\&
  \in \cH^\perp.
\end{align*}
where \annotate.
Next, we can show that the solution $\mH_k$ is unique. Indeed, by Theorem~2.10 of \citet{carlen2010trace}, the function $\<\mI,\phi(\mH)>$ is strictly convex, because the function $\phi(h)$ defined in \cref{eq:phi} is strictly convex.
Finally, we can prove \cref{eq:H_order}. It follows from the operator monotonicity of the function $h \mapsto -1/\sqrt{h}$, which is implied by L\"owner-Heinz Theorem \citep[Theorem~2.6]{carlen2010trace}, and the ordering $\proj[\cH]{\mS_{k+1}} \succeq \proj[\cH]{\mS_k}$, which is implied by \Cref{ass:H:proj} and the definition of $\mS_k$ in \cref{eq:H}.
\qed

\proofsubsection{lem:grad}

\Cref{ass:f} implies the following inequality for all $x \in \cX$ and $\nabla f(x) \in \partial f(x)$:
\begin{equation}
  f(x^*) \leq f(x) + \<\nabla f(x),x^* - x> + \tfrac{1}{1+\nu}\normt{\mL}^{\frac{1-\nu}{2}}\norm{x^* - x}^{1+\nu}_{\mL}.
\end{equation}
In the case $\nu \in (0,1]$, we can minimize the right-hand side in $x$, which gives the following:
\begin{equation}
  \norm{\nabla f(x)}_{\mL^{-1}}^{\frac{1+\nu}{\nu}} \leq \left(\tfrac{1+\nu}{\nu}\right)\normt{\mL}^{\frac{1-\nu}{2\nu}}\left(f(x) - f(x^*)\right).
\end{equation}
Taking both sides in the power $\frac{2\nu}{1+\nu}$ gives the desired inequality in the case $\nu \in (0,1]$. Finally, in the case $\nu = 0$, minimizing the right-hand side of the previous upper bound on $f(x^*)$ gives the following:
\begin{equation}
  f(x^*) \leq f(x) +
  \begin{cases}
    0 & \norm{\nabla f(x)}_{\mL^{-1}}^2 \leq \normt{\mL}\\
    -\infty & \norm{\nabla f(x)}_{\mL^{-1}}^2 > \normt{\mL}
  \end{cases}.
\end{equation}
It remains to use the fact that both $f(x)$ and $f(x^*)$ are finite to obtain the desired inequality in the case $\nu=0$.\qed

\proofsubsection{lem:dist}

\begin{enumerate}[label={\bf(\roman*)}]
  \item {\bf Non-negativity.} It is obvious.
  \item {\bf Absolute homogenity.} For arbitrary $t \in \R$ we can obtain the following:
    \begin{align*}
      \dist{t x}
      \at{use the definition of $\dist{x}$ in \cref{eq:dist}}=
      \norms{\proj[\cH]{t^2\mX}}^{1/2}
      \at{uses the linearity of the projection onto $\cH$ and the absolute homogentiy of $\norms{}$}=
      \abs{t}\cdot\norms{\proj[\cH]{\mX}}^{1/2}
      \at{use the definition of $\dist{x}$ in \cref{eq:dist}}=
      \abs{t}\cdot\dist{x},
    \end{align*}
    where \annotate.
  \item {\bf Positive definiteness.} Let $\dist{x} = 0$. Then $\proj[\cH]{\mX} = 0$, which implies the following:
    \begin{align*}
      0
      =
      \<\mI,\proj[\cH]{\mX}>
      \at{uses the fact that $\mI \in \cH$ due to \Cref{ass:H:func}}=
      \<\mI,\mX>
      \at{uses the definition of $\mX$ in \cref{eq:dist}}=
      \sqn{x}
    \end{align*}
    where \annotate. Hence, we get $x = 0$.
  \item {\bf Subadditivity.} Let $x,y \in \cX$. Then we can obtain the following:
    \begin{align*}
      \dist{x+y}
      &\at{use the definition of $\dist{x}$ in \cref{eq:dist}}=
      \norms{\proj[\cH]{(x+y)\<x+y,>}}^{1/2}
      \\&\at{uses the bilinearity of the mapping $x \mapsto x\<x,>$ and an arbitrary constant $c \in \R$}=
      \norms{\proj[\cH]{(1+c^2)x\<x,> + (1+1/c^2)y\<y,> - (cx - y/c)\<cx - y/c,>}}^{1/2}
      \\&\at{uses \Cref{ass:H:proj}, the linearity of the projection onto $\cH$, and the fact that $\norms{}$ is order-preserving on $\Symp$}\leq
      \norms{(1+c^2)\proj[\cH]{x\<x,>} + (1+1/c^2)\proj[\cH]{y\<y,>}}^{1/2}
      \\&\at{uses the subadditivity and absolute homogenity of $\norms{}$}\leq
      ((1+c^2)\norms{\proj[\cH]{x\<x,>}} + (1+1/c^2)\norms{\proj[\cH]{y\<y,>}})^{1/2}
      \\&\at{can be obtain by minimizing in $c$}=
      \norms{\proj[\cH]{x\<x,>}}^{1/2} + \norms{\proj[\cH]{y\<y,>}}^{1/2}
      \\&\at{use the definition of $\dist{x}$ in \cref{eq:dist}}=
      \dist{x} + \dist{y}.
    \end{align*}
    where \annotate.
\end{enumerate}
The proof is now complete.
\qed

\newpage

\section{Proofs for \Cref{sec:alg}}

\proofsubsection{lem:f_bound}

Let $r_k = x_k - x^*$ and $\mR_k = r_k\<r_k,>$. We can rewrite $\tfrac{1}{2}\sqn{r_{k+1}}_{\mH_k^{-1}}$ as follows:
\begin{equation}\label{eq:r_k}
  \tfrac{1}{2}\sqn{r_{k+1}}_{\mH_k^{-1}}
  \at{uses \cref{eq:gd}}=
  \tfrac{1}{2}\sqn{r_k}_{\mH_k^{-1}}
  -\<g_k,r_k> + \tfrac{1}{2}\sqn{g_k}_{\mH_k},
\end{equation}
where \annotate.
Next, we sum these equations for $k=0\clc K$ and get the following:
\begin{align*}
  \mi[1]\tsum_{k=0}^{K}\<g_k,r_k>
  \\&=
  \tfrac{1}{2}\tsum_{k=0}^{K}\sqn{g_k}_{\mH_k}
  +\tfrac{1}{2}\tsum_{k=0}^{K}\left(\sqn{r_k}_{\mH_k^{-1}} -\sqn{r_{k+1}}_{\mH_k^{-1}}\right)
  \\&=
  \tfrac{1}{2}\tsum_{k=0}^{K}\sqn{g_k}_{\mH_k}
  +\tfrac{1}{2}\sqn{r_0}_{\mH_0^{-1}}
  +\tfrac{1}{2}\tsum_{k=1}^{K}\sqn{r_k}_{\mH_k^{-1} - \mH_{k-1}^{-1}}
  -\tfrac{1}{2}\sqn{r_{K+1}}_{\mH_{K+1}^{-1}}
  \\&\leq
  \tfrac{1}{2}\tsum_{k=0}^{K}\sqn{g_k}_{\mH_k}
  +\tfrac{1}{2}\<\mR_0,\mH_0^{-1}>
  +\tfrac{1}{2}\tsum_{k=1}^{K}\<\mR_k,\mH_{k}^{-1}-\mH_{k-1}^{-1}>
  \\&\at{use the properties of the projection and the fact that $\mH_k^{-1} \in \cH$ due to \Cref{ass:H:func} and \cref{eq:H_exp}}=
  \tfrac{1}{2}\tsum_{k=0}^{K}\sqn{g_k}_{\mH_k}
  +\tfrac{1}{2}\<\proj[\cH]{\mR_0},\mH_0^{-1}>
  +\tfrac{1}{2}\tsum_{k=1}^{K}\<\proj[\cH]{\mR_k},\mH_{k}^{-1}-\mH_{k-1}^{-1}>
  \\&\at{uses the H\"older's inequality for Schatten norms, the definition of the norm $\dist{}$ in \cref{eq:dist}, and the inequality in \cref{eq:R}}\leq
  \tfrac{1}{2}\tsum_{k=0}^{K}\sqn{g_k}_{\mH_k}
  +\tfrac{1}{2}\cR^2\normt{\mH_0^{-1}}
  +\tfrac{1}{2}\cR^2\tsum_{k=1}^{K}\normt{\mH_{k}^{-1}-\mH_{k-1}^{-1}}
  \\&\at{uses the fact that $\mH_{k+1}^{-1}\succeq \mH_k^{-1}$, which is implied by \cref{eq:H_order} and the operator monotonicity of the function $h \mapsto -1/h$, which is implied by L\"owner-Heinz Theorem \citep[Theorem~2.6]{carlen2010trace}}=
  \tfrac{1}{2}\tsum_{k=0}^{K}\sqn{g_k}_{\mH_k}
  +\tfrac{1}{2}\cR^2\<\mI,\mH_0^{-1}>
  +\tfrac{1}{2}\cR^2\tsum_{k=1}^{K}\<\mI,\mH_{k}^{-1}-\mH_{k-1}^{-1}>
  \\&\at{uses \Cref{lem:ftl_btl}}\leq
  \tfrac{1}{2}\<\mH_{K},\mS_{K}>
  +\tfrac{1}{2}\<\mI, \phi(\mH_{K})>
  +\tfrac{1}{2}\cR^2\<\mI,\mH_{K}^{-1}>
  \\&\at{use the fact that $\mH_k^{-1} \in \cH$ due to \Cref{ass:H:func} and \cref{eq:H_exp}}=
  \tfrac{1}{2}\<\mH_{K},\proj[\cH]{\mS_{K}}>
  +\tfrac{1}{2}\<\mI, \phi(\mH_{K})>
  +\tfrac{1}{2}\cR^2\<\mI,\mH_{K}^{-1}>
\end{align*}
where \annotate.

Next, using the definition of the potential function $\phi(\mH)$ in \cref{eq:phi}, the expression for $\mH_k$ in \cref{eq:H_exp}, and the definition $\eta = \cR$, we get the following inequality:
\begin{align*}
  \tsum_{k=0}^{K}\<g_k,r_k>
  &\leq
  \tfrac{1}{2}\<\mH_K,\delta \mI + \proj[\cH]{\mS_{K}}>
  +\tfrac{1}{2}(\eta^2 + \cR^2)\<\mI,\mH_{K}^{-1}>
  \\&\at{uses the definition $\eta=\cR$}=
  \tfrac{3}{2}\cR\<\mI,(\delta \mI + \proj[\cH]{\mS_{K}})^{1/2}>,
\end{align*}
where \annotate.
After taking the expectation, recalling that $\xi_k$ is independent of $x_k$, and using \Cref{ass:sigma:mean}, we get
\begin{align*}
  \tsum_{k=0}^{K}\E{\<\nabla f(x_k),r_k>}
  &\leq
  \tfrac{3}{2}\cR\E{\<\mI,(\delta \mI + \proj[\cH]{\mS_{K}})^{1/2}>}
  \\&\at{uses the concavity of the function $\mH \mapsto \<\mI,\mH^{1/2}>$, which is implied by Theorem~2.10 of \citet{carlen2010trace}, and the linearity of the projection onto $\cH$}\leq
  \tfrac{3}{2}\cR\<\mI,(\delta \mI + \proj[\cH]{\E{\mS_{K}}})^{1/2}>
  \\&\at{uses the fact that function $\mH \mapsto \<\mI,\mH^{1/2}>$ is subadditive for $\mH \in \Symp$, which is implied by Lemma~3 of \citet{an2025asgo}}\leq
  \tfrac{3}{2}\cR\<\mI,\proj[\cH]{\E{\mS_{K}}}^{1/2}>
  +\tfrac{3}{2}\sqrt{\delta}\cR\normt{\mI}
  \\&=
  \tfrac{3}{2}\cR\<\mI,\proj[\cH]{\E{\mS_{K}}}^{1/2}>
  +\tfrac{3}{2}\sqrt{\delta}\cR\dim{\cX}
\end{align*}
where \annotate.
It remains to use the convexity property from \Cref{ass:f}.
\qed

\proofsubsection{lem:S_bound}

Let $\mG_k,\mN_k \in \Sympp$ be defined as follows:
\begin{equation}
  \mG_k = \tsum_{i=0}^{k}\nabla f(x_k)\<\nabla f(x_k),>,
  \quad
  \mN_k = \tsum_{i=0}^{k}n(x_k;\xi_k)\<n(x_k;\xi_k),>.
\end{equation}
Then, we can obtain the following:
\begin{align*}
  \E{\mS_k}
  &\at{uses the definition of $\mS_k$ in \cref{eq:H} and \Cref{ass:sigma}}=
  \tsum_{i=0}^{k} \E{(\nabla f(x_k) + n(x_k;\xi_k))\<\nabla f(x_k) + n(x_k;\xi_k),>}
  \\&=
  \E{\mG_k + \mN_k}
  +\tsum_{i=0}^{k} \E{\nabla f(x_k)\<n(x_k;\xi_k),> + n(x_k;\xi_k)\<\nabla f(x_k),>}
  \\&\at{uses \Cref{ass:sigma:mean} and the fact that $\xi_k$ is independent of $x_k$}=
  \E{\mG_k + \mN_k},
\end{align*}
where \annotate.
Using this, we obtain the following relation:
\begin{align*}
  \<\mI,\proj[\cH]{\E{\mS_k}}^{1/2}>
  &=
  \<\mI,\proj[\cH]{\E{\mG_k + \mN_k}}^{1/2}>
  \\&\at{uses the linearity of the expectation and the projection onto $\cH$}=
  \<\mI,\left[\proj[\cH]{\E{\mG_k}} + \proj[\cH]{\E{\mN_k}}\right]^{1/2}>
  \\&\at{uses the fact that function $\mH \mapsto \<\mI,\mH^{1/2}>$ is subadditive for $\mH \in \Symp$, which is implied by Lemma~3 of \citet{an2025asgo}}\leq
  \<\mI,\proj[\cH]{\E{\mG_k}}^{1/2}>
  +\<\mI,\proj[\cH]{\E{\mN_k}}^{1/2}>,
\end{align*}
where \annotate.

We can upper-bound $\<\mI,\proj[\cH]{\E{\mN_k}}^{1/2}>$ as follows:
\begin{align*}
  \<\mI,\proj[\cH]{\E{\mN_k}}^{1/2}>
  &=
  \<\mSigma^{1/2},\mSigma^{-1/2}\proj[\cH]{\E{\mN_k}}^{1/2}>
  \\&\at{uses the Cauchy-Schwarz inequality}\leq
  \norm{\mSigma^{1/2}}\norm{\mSigma^{-1/2}\proj[\cH]{\E{\mN_k}}^{1/2}}
  \\&\at{uses the definition of $\norm{}$ and $\normt{}$}=
  \sqrt{\normt{\mSigma}\<\mSigma^{-1},\proj[\cH]{\E{\mN_k}}>}
  \\&\at{uses the linearity of the expectation and the fact that $\mSigma^{-1} \in \cH$, which is implied by \Cref{ass:H:func,ass:sigma:variance}}=
  \sqrt{\normt{\mSigma}\E{\<\mSigma^{-1},\mN_k>}}
  \\&\at{uses the definition of $\mN_k$}\leq
  \sqrt{\normt{\mSigma}\tsum_{i=0}^k\E{\sqn{n(x_i;\xi_i)}_{\mSigma^{-1}}}}
  \\&\at{uses \Cref{ass:sigma:variance}}\leq
  \sqrt{k+1}\normt{\mSigma}
\end{align*}
where \annotate.

Similarly, we can upper-bound $\<\mI,\proj[\cH]{\E{\mG_k}}^{1/2}>$ as follows:
\begin{align*}
  \<\mI,\proj[\cH]{\E{\mG_k}}^{1/2}>
  &\at{uses steps similar to the above calculations}\leq
  \sqrt{\normt{\mL}\<\mL^{-1},\proj[\cH]{\E{\mG_k}}>}
  \\&\at{uses the linearity of the expectation and the fact that $\mL^{-1} \in \cH$, which is implied by \Cref{ass:H:func} and \Cref{ass:f}}=
  \sqrt{\normt{\mL}\E{\<\mL^{-1},\mG_k>}}
  \\&\at{uses the definition of $\mG_k$}\leq
  \sqrt{\normt{\mL}\tsum_{i=0}^k\E{\sqn{\nabla f(x_i)}_{\mL^{-1}}}}
  \\&\at{uses \Cref{lem:grad}}\leq
  \normt{\mL}^{\frac{1}{1+\nu}}\sqrt{\tsum_{i=0}^k
  \E*{\left[f(x_i) - f(x^*)\right]^{\frac{2\nu}{1+\nu}}}}
  \\&\at{use the concavity of the function $t \mapsto t^{\frac{2\nu}{1+\nu}}$ for $\nu \in [0,1]$}\leq
  \normt{\mL}^{\frac{1}{1+\nu}}\sqrt{\tsum_{i=0}^k
  \left[\E{f(x_i) - f(x^*)}\right]^{\frac{2\nu}{1+\nu}}}
  \\&\at{use the concavity of the function $t \mapsto t^{\frac{2\nu}{1+\nu}}$ for $\nu \in [0,1]$}\leq
  \normt{\mL}^{\frac{1}{1+\nu}}\sqrt{(k+1)^{\tfrac{1-\nu}{1+\nu}}
  \left[\tsum_{i=0}^k\E{f(x_i) - f(x^*)}\right]^{\frac{2\nu}{1+\nu}}}
  \\&=
  \sqrt{k+1}^{\tfrac{1-\nu}{1+\nu}}\normt{\mL}^{\frac{1}{1+\nu}}
  \left[\tsum_{i=0}^k\E{f(x_i) - f(x^*)}\right]^{\frac{\nu}{1+\nu}},
\end{align*}
where \annotate.
\qed

\proofsubsection{thm:alg}

Using \Cref{lem:f_bound,lem:S_bound}, we get the following inequality:
\begin{align*}
  \tsum_{k=0}^K\E{f(x_k) - f(x^*)}
  &\leq
  \tfrac{3}{2}\sqrt{K+1}^{\tfrac{1-\nu}{1+\nu}}\cR\normt{\mL}^{\frac{1}{1+\nu}}
  \left[\tsum_{k=0}^K\E{f(x_k) - f(x^*)}\right]^{\frac{\nu}{1+\nu}}
  \\&
  +\tfrac{3}{2}\sqrt{K+1}\cR\normt{\mSigma}
  +\tfrac{3}{2}\sqrt{\delta}\cR\dim{\cX},
\end{align*}
which implies the following inequality:
\begin{align*}
  \tsum_{k=0}^K\E{f(x_k) - f(x^*)}
  &\leq
  3\sqrt{K+1}^{1-\nu}\normt{\mL}\cR^{1+\nu}
  \\&
  +3\sqrt{K+1}\normt{\mSigma}\cR
  +3\sqrt{\delta}\cR\dim{\cX}.
\end{align*}
It remains to use the convexity property in \Cref{ass:f} and the definition of $\ox_K$ on \cref{line:ox} of \Cref{alg}.
\qed

\newpage

\section{Proofs for \Cref{sec:alg2}}

\proofsubsection{lem:alpha}

We can upper-bound $\tsum_{k=0}^K \E{f_k(x^*) - f_k(x_{k+1})}$ as follows:
\begin{align*}
  \mi[3]\tsum_{k=0}^K \E{f_k(x^*) - f_k(x_{k+1})}
  \\&\at{uses the definition of the functions $f_k(x)$ in \cref{eq:f_k}}=
  \tsum_{k=0}^K \alpha_k^{-2}\E{f(\alpha_k x^* + (1-\alpha_k)\ox_k) - f(\alpha_k x_{k+1} + (1-\alpha_k)\ox_k)}
  \\&\at{uses the definition of $\ox_{k+1}$ on \cref{line2:ox} of \Cref{alg2} and the convexity property in \Cref{ass:f}}\leq
  \tsum_{k=0}^K \alpha_k^{-2}\E{\alpha_k f(x^*) + (1-\alpha_k)f(\ox_k) - f(\ox_{k+1})}
  \\&=
  \alpha_K^{-2}\E{f(x^*) - f(x_{K+1})}
  +\alpha_0^{-2}(1-\alpha_0)\E{f(x^*) - f(\ox_0)}
  \\&
  +\tsum_{k=1}^{K} (\alpha_k^{-2}(1-\alpha_k) - \alpha_{k-1}^{-2})\E{f(\ox_k) - f(x^*)}
  \\&\at{uses the definition $\alpha_k = 2/(k+2)$}\leq
  \tfrac{1}{4}(K+2)^2\E{f(x^*) - f(x_{K+1})},
\end{align*}
where \annotate.
\qed

\proofsubsection{lem:LH}

Let $\mB = \mL$ (the case $\mB = \mSigma$ is analogous).
Let $\cA_k(\mQ)\colon \Sympp \to \R$ be the objective function in \cref{eq:LH}:
\begin{equation}
  \cA_k(\mQ) = \<\mQ,\mL\mS_k> + \<\mL,\delta\mQ - \eta^2 \ln(\mQ)>.
\end{equation}
From \Cref{ass:H:func}, it follows that $\mH_k^2 \in \cH \cap \Sympp$. In addition, from the L\"owner-Heinz Theorem \citep[Theorem~2.6]{carlen2010trace}, it follows that the function $\cA_k(\mQ)$ is convex. Hence, it remains to prove that the first-order stationarity condition holds, that is, the differential of $\cA_k(\mQ)$ is zero on $\cH$ at $\mH_k^2$:
\begin{equation}
  \d \cA_k(\mH_k^2)[\mH] = 0
  \;\;\text{for all}\;\;\mH \in \cH.
\end{equation}
The following \Cref{lem:differential} will be used to compute the differential $\d \cA_k(\mQ)[\mH]$.
\begin{lemma}<lem:differential>
  Under \Cref{ass:LH}, let the function $\cB(\mQ)\colon \Sympp \to \R$ be defined as follows:
  \begin{equation}
    \cB(\mQ) = \<\mL,\ln(\mQ)>.
  \end{equation}
  Then the differential of the function $\cB(\mQ)$ for all $\mQ \in \cH \cap \Sympp$ is given as follows:
  \begin{equation}
    \d\cB(\mQ)[\mH] = \<\mL\mQ^{-1},\mH> \;\;\text{for all}\;\;\mH \in \cH.
  \end{equation}
\end{lemma}
Using \Cref{lem:differential}, we can compute the differential $\d \cA_k(\mH_k^2)[\mH]$ for $\mH \in \cH$ as follows:
\begin{align*}
  \d \cA_k(\mH_k^2)[\mH]
  &\at{uses \Cref{lem:differential}}=
  \<\mL(\mS_k + \delta\mI - \eta^2 \mH_k^{-2}),\mH>
  \\&\at{uses \cref{eq:H_exp}}=
  \<\mL(\mS_k -\proj[\cH]{\mS_k}),\mH>
  \\&\at{uses \Cref{ass:LH}}=
  \<(\mS_k -\proj[\cH]{\mS_k}),\mL\mH>
  \\&\at{uses the fact that $\mL\mH \in \cH$, which is implied by the following \Cref{lem:commute}}=
  \<(\mS_k -\proj[\cH]{\mS_k}),\mL\mH>
\end{align*}
where \annotate.
\begin{lemma}<lem:commute>
  Under \Cref{ass:LH}, $\mL\mH \in \cH$ for all $\mH \in \cH$.
\end{lemma}
The proof is now complete.
\qed

\proofsubsubsection{lem:differential}

Let constants $a,b \in \R$ be chosen to satisfy the following inequalities:
\begin{equation}
  \mO \prec a\mI \prec \mQ \prec b\mI.
\end{equation}
Let $\mH \in \cH$ such that $\norms{\mH-\mQ}\leq \min\left\{(\lmin(\mQ) - a),b-\lmax(\mQ)\right\}$. Hence, it is easy to verify that the following inequalities hold:
\begin{equation}
  a\mI \preceq \mQ + \mH \preceq b\mI.
\end{equation}
Next, we fix an arbitrary $\epsilon > 0$. By the Weierstrass approximation theorem, there exists a polynomial $p_n(t) = \sum_{i=0}^{n}c_i t^i$ such that $p_n(a) = \ln(a)$, $p_n'(a) = 1/a$, and whose second derivative approximates the function $t\mapsto -1/t^2$ on the segment $[a,b]$ up to the precision $\epsilon$:
\begin{equation}
  \abs{p_n''(t) + 1/t^2} \leq \epsilon
  \;\;\text{for all}\;\; t\in [a,b].
\end{equation}
From this, using the standard integration arguments, we can conclude that the following approximation inequalities hold for all $t \in [a,b]$:
\begin{equation}
  \abs{p_n'(t) - 1/t}\leq \epsilon(b-a),\qquad
  \abs{p_n(t) - \ln(t)}\leq \tfrac{1}{2}\epsilon(b-a)^2.
\end{equation}
Further, we obtain the following:
\begin{align*}
  \mi[3]\abs{\cB(\mQ+\mH) - \cB(\mQ) - \tint_{0}^{1}\<\mL(\mQ + \tau \mH)^{-1},\mH>\d\tau}
  \\&\at{uses \Cref{def:op_f}, the approximation inequalities above, and the H\"older's inequality for Schatten norms}\leq
  \abs{\<\mL,p_n(\mQ+\mH) - p_n(\mQ)> - \tint_{0}^{1}\<\mL p_n'(\mQ+\tau\mH),\mH>\d\tau}
  \\&
  +\normt{\mL}\cdot\left(\tfrac{1}{2}\epsilon(b-a)^2 + \tfrac{1}{2}\epsilon(b-a)^2\right) + \normt{\mL\mH}\cdot\epsilon(b-a)
  \\&=
  \abs{\<\mL,p_n(\mQ+\mH) - p_n(\mQ)> - \tint_{0}^{1}\<\mL p_n'(\mQ+\tau\mH),\mH>\d\tau}
  +\epsilon \left(b^2\normt{\mL} + b\normt{\mL\mH}\right)
  \\&\at{Uses the fact that $p_n(t)$ is a polynomial and the fact that $\mQ\mL = \mL\mQ$ and $\mH\mL = \mL\mH$ due to \Cref{ass:LH}}=
  \epsilon \left(b^2\normt{\mL} + b\normt{\mL\mH}\right).
\end{align*}
where \annotate. Next, we take the limit $\epsilon \to 0$ and use the fundamental theorem of calculus and the continuity of the map $\mQ \mapsto \mQ^{-1}$ on $\Sympp$, which implies the following:
\begin{equation}
  \tfrac{\d}{\d\tau}\cB(\mQ + \tau \mH)|_{\tau = 0} = \<\mL\mQ^{-1},\mH>.
\end{equation}
Since the right-hand side is continuous in $\mQ$, we can conclude that the function $\cB(\mQ)$ is differentiable and its differential is equal to the right-hand side.

\proofsubsubsection{lem:commute}
Since the operators $\mL$ and $\mH$ are self-adjoint and commute, they are simultaneously diagonalizeable:
\begin{equation*}
  \mL = \tsum_i \lambda_i\cdot u_i\<u_i,>
  \quad\text{and}\quad
  \mH = \tsum_i \mu_i\cdot u_i\<u_i,>,
\end{equation*}
where $\lambda_i$ and $\mu_i$ are the (possibly repeating) eigenvalues of the operators $\mL$ and $\mH$, respectively, $\{u_i\}\subset \cX$ is an orthonormal basis of the common eigenvectors in the space $\cX$. Hence, the operator $\mL\mH$ is also diagonalizeable as follows:
\begin{equation*}
  \mL\mH = \tsum_i \lambda_i\mu_i \cdot u_i\<u_i,>.
\end{equation*}
Further, let $I_{\lambda} = \set{i\given \lambda_i = \lambda}$ and $J_{\mu} = \set{j \given \mu_j = \mu}$ for arbitrary $\lambda,\mu \in \R$. Let $p(t)$ be a polynomial such that $p(\lambda) = 1$ and $p(\lambda_i) = 0$ for $i \notin I$. Using \Cref{ass:H:func}, we can conclude that
\begin{equation*}
  p(\mL) = \tsum_{i \in I_{\lambda}} u_i\<u_i,> \in \cH.
\end{equation*}
Similarly, by constructing a polynomial $q(t)$such that $q(\mu) = 1$ and $q(\mu_j) = 0$ for $j \notin J$ and using \Cref{ass:H:func}, we can show that the following inclusion holds:
\begin{equation*}
  q(\mH) = \tsum_{j \in J_{\mu}} u_j\<u_j,> \in \cH.
\end{equation*}
Hence, using \Cref{ass:H:func}, we obtain the following inclusion:
\begin{equation*}
  p(\mL) + q(\mH) = \tsum_{i \in I_{\lambda} \triangle J_{\mu}} u_i\<u_i,> + 2\tsum_{i \in I_{\lambda} \cap J_{\mu}} u_i\<u_i,>.
\end{equation*}
Finally, we can construct a polynomial $s(t)$ such $s(2) = 1$ and $s(1) = 0$.
Using \Cref{ass:H:func}, we can show that
\begin{equation*}
  s(p(\mL) + q(\mH)) = \tsum_{i \in I_{\lambda} \cap J_{\mu}} u_i\<u_i,> \in \cH.
\end{equation*}
From this fact and the above eigendecomposition of the operator $\mL\mH$, it follows that $\mL\mH \in \cH$.
\qed

\proofsubsection{lem:ftl_btl2}

Let $\mB = \mL$ (the case $\mB = \mSigma$ is analogous).
Let functions $l_{-1}(\mQ)\clc l_k(\mQ)\colon \Sympp\cap\cH \to \R$ be defined as follows:
\begin{equation}
  l_{-1}(\mQ) = \<\mL,\delta\mQ - \eta^2 \ln(\mQ)>,\quad
  l_i(\mQ) = \sqn{g_i}_{\mQ\mL} \text{\;\;for\;\;}i=0\clc k.
\end{equation}
Let the operators $\mQ_{-1}\clc\mQ_k\in\cH\cap \Sympp$ be defined as follows:
\begin{equation}
  \mQ_{-1} = (\eta^2/\delta)\mI,\quad
  \mQ_i = \mH_i^2 \text{\;\;for\;\;}i=0\clc k.
\end{equation}
Using \Cref{lem:LH}, we can show that the following relation holds for all $i = -1\clc k$:
\begin{equation*}
  \mQ_i = \argmin_{\mQ \in \cH \cap \Sympp} \tsum_{i=-1}^{k} l_i(\mQ).
\end{equation*}
Next, we get the following inequality:
\begin{align*}
  \tsum_{i=0}^{k} \sqn{g_i}_{\mL\mH_i^2}
  &\at{use the definition of the functions $l_i(\mH)$, the definition of the operators $\mQ_i$}=
  \tsum_{i=0}^{k} l_i(\mQ_i)
  \\&=
  \tsum_{i=-1}^{k} l_i(\mQ_i) - l_{-1}(\mQ_{-1})
  \\&\at{uses \cref{eq:ftl_btl}}\leq
  \tsum_{i=-1}^{k} l_i(\mQ_k) - l_{-1}(\mQ_{-1})
  \\&\at{use the definition of the functions $l_i(\mH)$, the definition of the operators $\mQ_i$}=
  \<\mL\mH_k^2,\delta \mI + \mS_k> - \eta^2\<\mL,  \ln(\mH_k^2)> - \eta^2\<\mL,\mI> + \eta^2\<\mL, \ln((\eta^2/\delta)\mI)>
  \\&\at{uses \Cref{lem:commute}, \Cref{ass:H:func}, and the properties of the projection onto $\cH$}=
  \<\mL\mH_k^2,\delta \mI + \proj[\cH]{\mS_k}> - \eta^2\<\mL,  \ln(\mH_k^2)>
  - \eta^2\normt{\mL} + \eta^2\<\mL, \ln(\eta^2\mI/\delta)>
  \\&\at{use \cref{eq:H_exp} and \Cref{def:op_f}}=
  \eta^2\normt{\mL} - \eta^2\<\mL,  \ln(\delta\mH_k^2/\eta^2)>
  - \eta^2\normt{\mL}
  \\&\at{use \cref{eq:H_exp} and \Cref{def:op_f}}=
  \eta^2\<\mL, \ln(\delta \mI + \proj[\cH]{\mS_k})> + \eta^2\normt{\mL}\ln\tfrac{1}{\delta}
  \\&\at{uses the H\"older's inequality for Schatten norms}\leq
  \eta^2\normt{\mL}\ln(\norms{\delta \mI + \proj[\cH]{\mS_k}}) + \eta^2\normt{\mL}\ln\tfrac{1}{\delta}
  \\&\at{uses the inequality $\norms{} \leq \normt{}$}\leq
  \eta^2\normt{\mL}\ln(\sqnt{(\delta \mI + \proj[\cH]{\mS_k})^{1/2}}) + \eta^2\normt{\mL}\ln\tfrac{1}{\delta}
  \\&\at{uses \cref{eq:H_exp}}=
  \eta^2\normt{\mL}\ln\left(\tfrac{1}{\delta}\eta^2\sqnt{\mH_k^{-1}}\right),
\end{align*}
where \annotate. It remains to take the expectation and use the concavity of the function $t \mapsto \ln(t^2)$ and the Jensen's inequality.
\qed

\proofsubsection{lem:f_bound2}

Let $n_k = g_k - \nabla f_k(x_k)$ and $r_k = x_k - x^*$.
We can obtain the following inequality:
\begin{align*}
  \E{f_k(x_{k+1})}
  &\at{uses the definition of the function $f_k(x)$ in \cref{eq:f_k} and \Cref{ass:f}}\leq
  \E*{f_k(x_k) + \<\nabla f_k(x_k),x_{k+1} - x_k> + \tfrac{1}{1+\nu}\alpha_k^{\nu-1}\normt{\mL}^{\frac{1-\nu}{2}}\norm{x_{k+1}-x_k}_\mL^{1+\nu}}
  \\&\at{uses \cref{eq:gd} and \Cref{ass:LH}}=
  \E*{f_k(x_k) - \<\nabla f_k(x_k),\mH_k g_k> + \tfrac{1}{1+\nu}\alpha_k^{\nu-1}\normt{\mL}^{\frac{1-\nu}{2}}\norm{g_k}_{\mL\mH_k^2}^{1+\nu}}
  \\&\at{uses the definition of $n_k$}=
  \E*{f_k(x_k) - \sqn{g_k}_{\mH_k} + \<n_k,\mH_k g_k> + \tfrac{1}{1+\nu}\alpha_k^{\nu-1}\normt{\mL}^{\frac{1-\nu}{2}}\norm{g_k}_{\mL\mH_k^2}^{1+\nu}}
\end{align*}
where \annotate.
Next, similar to the proof of \Cref{lem:f_bound}, we can obtain the following inequality:
\begin{equation}
  \E{\tsum_{k=0}^K \<g_k,r_k>}
  \leq
  \E{\tfrac{1}{2}\tsum_{k=0}^{K}\sqn{g_k}_{\mH_k}
  +\tfrac{1}{2}\cR^2\<\mI,\mH_K^{-1}>},
\end{equation}
Combining this with the previous inequality gives the following:
\begin{align*}
  \mi[2]\tsum_{k=0}^K \E{f_k(x_{k+1}) - f_k(x^*)}
  \\&\leq
  \tfrac{1}{2}\cR^2\<\mI,\E{\mH_K^{-1}}>
  +\tsum_{k=0}^{K}\E*{
    \<n_k,\mH_k g_k>
    -\tfrac{1}{2}\sqn{g_k}_{\mH_k}
    +\tfrac{1}{1+\nu}\alpha_k^{\nu-1}\normt{\mL}^{\frac{1-\nu}{2}}\norm{g_k}_{\mL\mH_k^2}^{1+\nu}
  }
  \\&\at{uses the Young's inequality, \Cref{ass:LH}, and an arbitrary constant $c > 0$}\leq
  \tfrac{1}{2}\cR^2\<\mI,\E{\mH_K^{-1}}>
  +\tsum_{k=0}^{K}\E*{-\tfrac{1}{2}\sqn{g_k}_{\mH_k}
  +\tfrac{1}{1+\nu}\alpha_k^{\nu-1}\normt{\mL}^{\frac{1-\nu}{2}}\norm{g_k}_{\mL\mH_k^2}^{1+\nu}}
  \\&
  +\tsum_{k=0}^{K}\E*{
    \tfrac{c}{2}\sqn{g_k}_{\mSigma\mH_k^2}
    +\tfrac{1}{2c}\sqn{n_k}_{\mSigma^{-1}}
  }
  \\&\at{uses the definition of $\mS_k$ in \cref{eq:H}}=
  \tfrac{1}{2}\cR^2\<\mI,\E{\mH_K^{-1}}>
  +\tsum_{k=0}^{K}\E*{\tfrac{1}{2}\<\mS_{k-1} - \mS_k,\mH_k>
  +\tfrac{1}{1+\nu}\alpha_k^{\nu-1}\normt{\mL}^{\frac{1-\nu}{2}}\norm{g_k}_{\mL\mH_k^2}^{1+\nu}}
  \\&
  +\tsum_{k=0}^{K}\E*{
    \tfrac{c}{2}\sqn{g_k}_{\mSigma\mH_k^2}
    +\tfrac{1}{2c}\sqn{n_k}_{\mSigma^{-1}}
  }
  \\&\at{uses \cref{eq:H_order}}\leq
  \tfrac{1}{2}\cR^2\<\mI,\E{\mH_K^{-1}}>
  +\tfrac{1}{2}\tsum_{k=0}^{K}\E*{\<\mS_{k-1},\mH_{k-1}> - \<\mS_k,\mH_k>
  }
  \\&
  +\tsum_{k=0}^{K}\E*{
    \tfrac{1}{1+\nu}\alpha_k^{\nu-1}\normt{\mL}^{\frac{1-\nu}{2}}\norm{g_k}_{\mL\mH_k^2}^{1+\nu}
    +\tfrac{c}{2}\sqn{g_k}_{\mSigma\mH_k^2}
    +\tfrac{1}{2c}\sqn{n_k}_{\mSigma^{-1}}
  }
  \\&\at{uses the definition of $\mH_k$ in \cref{eq:H_exp}}=
  \tfrac{1}{2}(\cR^2 - \eta^2)\<\mI,\E{\mH_K^{-1}}>
  +\tfrac{1}{2}\sqrt{\delta}\eta\normt{\mI}
  \\&
  +\tsum_{k=0}^{K}\E*{
    \tfrac{1}{1+\nu}\alpha_k^{\nu-1}\normt{\mL}^{\frac{1-\nu}{2}}\norm{g_k}_{\mL\mH_k^2}^{1+\nu}
    +\tfrac{c}{2}\sqn{g_k}_{\mSigma\mH_k^2}
    +\tfrac{1}{2c}\sqn{n_k}_{\mSigma^{-1}}
  }
  \\&\at{uses the definition of $n_k$ above, \Cref{ass:sigma:variance}, and the definition of the function $f_k(x)$ in \cref{eq:f_k}}\leq
  \tfrac{1}{2}(\cR^2 - \eta^2)\<\mI,\E{\mH_K^{-1}}>
  +\tfrac{1}{2}\sqrt{\delta}\eta\normt{\mI}
  +\tfrac{1}{2c}\normt{\mSigma}\tsum_{k=0}^K (1/\alpha_k^{2})
  \\&
  +\tsum_{k=0}^{K}\E*{
    \tfrac{1}{1+\nu}\alpha_k^{\nu-1}\normt{\mL}^{\frac{1-\nu}{2}}\norm{g_k}_{\mL\mH_k^2}^{1+\nu}
    +\tfrac{c}{2}\sqn{g_k}_{\mSigma\mH_k^2}
  }
  \\&\at{uses the H\"older's inequality, the concavity of the function $t\mapsto t^{\frac{1+\nu}{2}}$, and the Jensen's inequality for the expectation}\leq
  \tfrac{1}{2}(\cR^2 - \eta^2)\<\mI,\E{\mH_K^{-1}}>
  +\tfrac{1}{2}\sqrt{\delta}\eta\normt{\mI}
  +\tfrac{1}{2c}\normt{\mSigma}\tsum_{k=0}^K (1/\alpha_k^{2})
  \\&
  +\tfrac{c}{2}\E*{\tsum_{k=0}^{K}
    \sqn{g_k}_{\mSigma\mH_k^2}
  }
  +\tfrac{1}{1+\nu}\left(\tsum_{i=0}^{K}1/\alpha_i^2\right)^{\tfrac{1-\nu}{2}}\normt{\mL}^{\frac{1-\nu}{2}}
  \left(\E*{\tsum_{k=0}^{K}\sqn{g_k}_{\mL\mH_k^2}}\right)^{\frac{1+\nu}{2}}
  \\&\at{can be obtained by minimizing in $c>0$}=
  \tfrac{1}{2}(\cR^2 - \eta^2)\<\mI,\E{\mH_K^{-1}}>
  +\tfrac{1}{2}\sqrt{\delta}\eta\normt{\mI}
  +\left(\tsum_{i=0}^K 1/\alpha_i^{2}\right)^{\frac{1}{2}}\normt{\mSigma}^{\frac{1}{2}}\left(\E*{
    \tsum_{k=0}^{K}\sqn{g_k}_{\mSigma\mH_k^2}}
  \right)^{\frac{1}{2}}
  \\&
  +\tfrac{1}{1+\nu}\left(\tsum_{i=0}^{K}1/\alpha_i^2\right)^{\tfrac{1-\nu}{2}}\normt{\mL}^{\frac{1-\nu}{2}}
  \left(\E*{\tsum_{k=0}^{K}\sqn{g_k}_{\mL\mH_k^2}}\right)^{\frac{1+\nu}{2}}
  \\&\at{uses the definition of $\normt{}$}=
  \tfrac{1}{2}(\cR^2 - \eta^2)\E{\normt{\mH_K^{-1}}}
  +\tfrac{1}{2}\sqrt{\delta}\eta\normt{\mI}
  +\left(\tsum_{i=0}^K 1/\alpha_i^{2}\right)^{\frac{1}{2}}\normt{\mSigma}^{\frac{1}{2}}\left(\E*{
    \tsum_{k=0}^{K}\sqn{g_k}_{\mSigma\mH_k^2}}
  \right)^{\frac{1}{2}}
  \\&
  +\tfrac{1}{1+\nu}\left(\tsum_{i=0}^{K}1/\alpha_i^2\right)^{\tfrac{1-\nu}{2}}\normt{\mL}^{\frac{1-\nu}{2}}
  \left(\E*{\tsum_{k=0}^{K}\sqn{g_k}_{\mL\mH_k^2}}\right)^{\frac{1+\nu}{2}},
\end{align*}
where \annotate.
Next, using \Cref{lem:ftl_btl2}, we obtain the following technical \Cref{lem:tech}.

\begin{lemma}<lem:tech>
  Under the conditions of \Cref{lem:f_bound2}, for $\mB = \mL$ or $\mB = \mSigma$, and for all $\gamma \in (0,1)$, the following inequality holds:
  \begin{equation}
    \begin{aligned}
      \mi[5]
      \left(\tsum_{i=0}^{k}1/\alpha_i^2\right)^{1-\gamma}\normt{\mB}^{1-\gamma}
      \left(\E*{\tsum_{i=0}^{k}\sqn{g_i}_{\mB\mH_i^2}}\right)^{\gamma}
      \\&\leq
      \tfrac{1}{8}\eta^2\E{\normt{\mH_k^{-1}}}
      +2^\gamma\left(\tsum_{i=0}^{k}1/\alpha_i^2\right)^{1-\gamma}\normt{\mB}
      \eta^{2\gamma}
      \ln^{\gamma}(c_k(\mB,\gamma)),
    \end{aligned}
  \end{equation}
  where the constant $c(\mB,\gamma) > 0$ is defined as follows:
  \begin{equation}
    c_k(\mB,\gamma)
    =
    \max\left\{
      \exp(1),\;
      2^{3+\gamma}
      \gamma^{\gamma}
      \left(\tsum_{i=0}^{k}1/\alpha_i^2\right)^{1-\gamma}
      \tfrac{1}{\sqrt{\delta}}
      \normt{\mB}
    \eta^{2\gamma-1}\right\}.
  \end{equation}
\end{lemma}
Further, using \Cref{lem:tech} and the fact that $\normt{\mI} = \dim{\cX}$, we obtain the following inequality:
\begin{align*}
  \mi[2]\tsum_{k=0}^K \E{f_k(x_{k+1}) - f_k(x^*)}
  \\&\leq
  \left(\tfrac{1}{2}\cR^2 - \tfrac{1}{4}\eta^2\right)\E{\normt{\mH_K^{-1}}}
  +2^{\frac{1+\nu}{2}}\left(\tsum_{i=0}^{K}1/\alpha_i^2\right)^{\frac{1-\nu}{2}}\normt{\mL}
  \eta^{1+\nu}
  \ln \left(c_K\big(\mL,\tfrac{1+\nu}{2}\big)\right)
  \\&
  +2^{\frac{1}{2}}\left(\tsum_{i=0}^{K}1/\alpha_i^2\right)^{\frac{1}{2}}\normt{\mSigma}
  \eta
  \ln \left(c_K\big(\mSigma,\tfrac{1}{2}\big)\right)
  +\tfrac{1}{2}\sqrt{\delta}\eta\dim{\cX}
  \\&\at{uses the definition $\eta = 2\cR$}\leq
  2^{\frac{3(1+\nu)}{2}}\left(\tsum_{i=0}^{K}1/\alpha_i^2\right)^{\frac{1-\nu}{2}}\normt{\mL}
  \cR^{1+\nu}\ln \left(c_K\big(\mL,\tfrac{1+\nu}{2}\big)\right)
  \\&
  +2^{\frac{3}{2}}\left(\tsum_{i=0}^{K}1/\alpha_i^2\right)^{\frac{1}{2}}\normt{\mSigma}
  \cR\ln \left(c_K\big(\mSigma,\tfrac{1}{2}\big)\right)
  +\sqrt{\delta}\cR\dim{\cX}
  \\&\at{uses the definition $\alpha_k = 2/(k+2)$}\leq
  2^{\frac{1+5\nu}{2}}\left(\tsum_{i=1}^{K+2}i^2\right)^{\frac{1-\nu}{2}}\normt{\mL}
  \cR^{1+\nu}\ln \left(c_K\big(\mL,\tfrac{1+\nu}{2}\big)\right)
  \\&
  +2^{\frac{1}{2}}\left(\tsum_{i=1}^{K+2}i^2\right)^{\frac{1}{2}}\normt{\mSigma}
  \cR\ln \left(c_K\big(\mSigma,\tfrac{1}{2}\big)\right)
  +\sqrt{\delta}\cR\dim{\cX}
  \\&\at{uses the fact that $\sum_{i=1}^{K+2}i^2 \leq \tfrac{1}{3}(K+3)^3$ and $\nu \leq 1$}\leq
  2^{\frac{1+5\nu}{2}}3^{\frac{\nu-1}{2}}(K+3)^{\frac{3(1-\nu)}{2}}\normt{\mL}
  \cR^{1+\nu}\ln \left(c_K\big(\mL,\tfrac{1+\nu}{2}\big)\right)
  \\&
  +2^{\frac{1}{2}}3^{-\frac{1}{2}}(K+3)^{\frac{3}{2}}\normt{\mSigma}
  \cR\ln \left(c_K\big(\mSigma,\tfrac{1}{2}\big)\right)
  +\sqrt{\delta}\cR\dim{\cX}
  \\&\leq
  8(K+2)^{\frac{3(1-\nu)}{2}}\normt{\mL}
  \cR^{1+\nu}\ln \left(c_K\big(\mL,\tfrac{1+\nu}{2}\big)\right)
  \\&
  +2(K+2)^{\frac{3}{2}}\normt{\mSigma}
  \cR\ln \left(c_K\big(\mSigma,\tfrac{1}{2}\big)\right)
  +\sqrt{\delta}\cR\dim{\cX}
\end{align*}
where \annotate. Finally, we define $\cC_K = 32\ln \left(\max\{c_K\big(\mL,\tfrac{1+\nu}{2}\big),c_K\big(\mSigma,\tfrac{1}{2}\big)\}\right)$ and verify that \cref{eq:C} holds.
\qed

\proofsubsubsection{lem:tech}
We start with the following inequality:
\begin{align*}
  \mi[2]
  \left(\tsum_{i=0}^{k}1/\alpha_i^2\right)^{1-\gamma}\normt{\mB}^{1-\gamma}
  \left(\E*{\tsum_{i=0}^{k}\sqn{g_i}_{\mB\mH_i^2}}\right)^{\gamma}
  \\&\at{uses \Cref{lem:ftl_btl2}}\leq
  \left(\tsum_{i=0}^{k}1/\alpha_i^2\right)^{1-\gamma}\normt{\mB}^{1-\gamma}
  \left(\eta^2\normt{\mB}\ln\left[\tfrac{1}{\delta}\eta^2\left(\E{\normt{\mH_k^{-1}}}\right)^2\right]\right)^{\gamma}
  \\&\at{uses an arbitrary constant $c>0$}=
  \left(\tsum_{i=0}^{k}1/\alpha_i^2\right)^{1-\gamma}\normt{\mB}
  \left(2\gamma\eta^2
    \ln\left[
      \left(\tfrac{\eta}{c\sqrt{\delta}}\right)^{\frac{1}{\gamma}}
      \left(\E{\normt{\mH_k^{-1}}}\right)^{\frac{1}{\gamma}}
    \right]
    +
    2\eta^2\ln(c)
  \right)^{\gamma}
  \\&\at{uses the subadditivity of the function $t\mapsto t^\gamma$}\leq
  \left(\tsum_{i=0}^{k}1/\alpha_i^2\right)^{1-\gamma}\normt{\mB}
  \left[
    \left(2\gamma\eta^2
      \ln\left[
        \left(\tfrac{\eta}{c\sqrt{\delta}}\right)^{\frac{1}{\gamma}}
        \left(\E{\normt{\mH_k^{-1}}}\right)^{\frac{1}{\gamma}}
      \right]
    \right)^{\gamma}
    +
    \left(
      2\eta^2\ln(c)
    \right)^{\gamma}
  \right]
  \\&\at{uses the inequality $\ln(t)\leq t$ for $t > 0$}\leq
  \left(\tsum_{i=0}^{k}1/\alpha_i^2\right)^{1-\gamma}\normt{\mB}
  \left[
    \left(2\gamma\eta^2\right)^{\gamma}
    \left(\tfrac{\eta}{c\sqrt{\delta}}\right)
    \E{\normt{\mH_k^{-1}}}
    +
    \left(
      2\eta^2\ln(c)
    \right)^{\gamma}
  \right]
\end{align*}
where \annotate.
Next, we choose the constant $c > 0$ as follows:
\begin{equation}
  c =
  \max\left\{
    \exp(1),\;
    2^{3+\gamma}
    \gamma^{\gamma}
    \left(\tsum_{i=0}^{k}1/\alpha_i^2\right)^{1-\gamma}
    \tfrac{1}{\sqrt{\delta}}
    \normt{\mB}
  \eta^{2\gamma-1}\right\},
\end{equation}
which implies the following inequality:
\begin{align*}
  \mi[5]
  \left(\tsum_{i=0}^{k}1/\alpha_i^2\right)^{1-\gamma}\normt{\mB}^{1-\gamma}
  \left(\E*{\tsum_{i=0}^{k}\sqn{g_i}_{\mB\mH_i^2}}\right)^{\gamma}
  \\&\leq
  \tfrac{1}{8}\eta^2\E{\normt{\mH_k^{-1}}}
  +2^{\gamma}\left(\tsum_{i=0}^{k}1/\alpha_i^2\right)^{1-\gamma}\normt{\mB}
  \eta^{2\gamma}
  \ln^{\gamma}(c)
  \\&\at{uses the fact that $\ln(c) \geq 1$}\leq
  \tfrac{1}{8}\eta^2\E{\normt{\mH_k^{-1}}}
  +2^\gamma\left(\tsum_{i=0}^{k}1/\alpha_i^2\right)^{1-\gamma}\normt{\mB}
  \eta^{2\gamma}
  \ln(c),
\end{align*}
where \annotate.
\qed

\end{document}